\documentclass[12pt]{article}

\usepackage{float}
\usepackage{ccaption}
\usepackage{scicite} 
\usepackage{times}
\usepackage{graphicx}
\usepackage[caption = false]{subfig} 
\usepackage{fancyhdr}
\usepackage{rotating}
\usepackage{caption}
\usepackage{tabularx}
\usepackage[T1]{fontenc}

\newenvironment{sciabstract}{ \begin{quote} \bf}{\end{quote}}
\title{An abstract model of nonrandom, non-Lamarckian mutation in evolution using a multivariate estimation-of-distribution algorithm \\ 
\vspace{4mm}} 
\author{Liudmyla Vasylenko,$^{1,2}$ ~Adi Livnat$^{1,2\ast}$\\
\\
\normalsize{$^{1}$Department of Evolutionary and Environmental Biology, University of Haifa, 3103301, Israel,}\\
\normalsize{$^{2}$Institute of Evolution, University of Haifa, 3103301, Israel}\\
\normalsize{$^\ast$ To whom correspondence should be addressed; E-mail: alivnat@univ.haifa.ac.il}
}

\date{}

\begin{document} 
\baselineskip24pt
\maketitle 
\thispagestyle{fancy}

\newpage
\cfoot{\thepage}

\begin{sciabstract}
Abstract: At the fundamental conceptual level, two alternatives have traditionally been considered for how mutations arise and how evolution happens: 1) random mutation and natural selection, and 2) Lamarckism. Recently, the theory of Interaction-based Evolution (IBE) has been proposed, according to which mutations are neither random nor Lamarckian, but are influenced by information accumulating internally in the genome over generations.  Based on the estimation-of-distribution algorithms framework, we present a simulation model that demonstrates nonrandom, non-Lamarckian mutation concretely while capturing indirectly several aspects of IBE: selection, recombination, and nonrandom, non-Lamarckian mutation interact in a complementary fashion; evolution is driven by the interaction of parsimony and fit; and random bits do not directly encode improvement but enable generalization by the manner in which they connect with the rest of the evolutionary process. Connections are drawn to Darwin's observations that changed conditions increase the rate of production of heritable variation; to the causes of bell-shaped distributions of traits and how these distributions respond to selection; and to computational learning theory, where analogizing evolution to learning in accord with IBE casts individuals as examples and places the learned hypothesis at the population level. The model highlights the importance of incorporating internal integration of information through heritable change in both evolutionary theory and evolutionary computation.
\end{sciabstract}

\section{Introduction}

For over a century, two alternatives at the fundamental conceptual level have dominated thinking about mutation origination in evolution: 1) random mutation, and 2) Lamarckism \cite{Morgan1903,Fisher1930,Dawkins1986,LenskiMittler1993,Futuyma1998,KooninWolf2009,Merlin2010}. ``Random mutation'' does not mean simply that mutation is nondeterministic, nor that the probabilities of mutation origination are uniform across the genome. Rather, it means that mutation is accidental: that the causes of any particular mutation are unrelated to its biological consequences \cite{LenskiMittler1993,Futuyma1998}. Lamarckism, instead, assumes that the organism can somehow sense its immediate environment and respond directly to it with beneficial heritable change \cite{Lamarck1809}. However, Lamarckism requires ``reverse engineering'' \cite{KooninWolf2009} of an unintuitive kind, similar to reversing a one-way function \cite{Haig2007,LivnatPapadimitriou2016alg}, and has been largely rejected \cite{LuriaDelbruck1943,LederbergLederberg1952,LenskiMittler1993,Futuyma1998,Merlin2010}.

The lack of conceptual alternatives at the fundamental level until recently can be further clarified. Suppose that a random mutation occurs in a DNA-repair protein, changes the average mutation rate across many loci, and is favored by selection. Although this could hypothetically allow the average mutation rate to evolve adaptively \cite{FeldmanLiberman1986,AltenbergFeldman1987,Lynch2010,HodgkinsonEyre-Walker2011}, it leaves the causes of any particular mutation accidental \cite{Futuyma1998}. As Lenski and Mittler asserted: ``A fundamental tenet of evolutionary biology is that mutations are random events,'' where this randomness means that ``the likelihood of any particular mutational event is independent of its specific value to the organism'' \cite{LenskiMittler1993}.

Recently, however, a new theory was proposed, called Interaction-based Evolution (IBE), according to which mutations are neither random nor Lamarckian \cite{Livnat2013,Livnat2017,LivnatMelamed2023}. This theory holds that individual mutations as well as individual epigenetic changes are not random because their probabilities of origination depend on complex information in a biologically meaningful manner. However, they are also not Lamarckian, because the information they depend on does not arrive directly from the immediate environment; instead, it accumulates in the genome over generations. At each generation, heritable changes arise based on the internal information accumulated thus far, and those that survive become a part of the information affecting future generations’ mutations. Accordingly, evolution is akin to a long-term learning process based on nonrandom, non-Lamarckian heritable changes at its core \cite{Livnat2013,Livnat2017,LivnatMelamed2023}.

Surprisingly, despite the fact that ``random mutation'' refers to the rates of individual mutations, and despite its centrality in evolutionary thinking \cite{LenskiMittler1993,Futuyma1998}, mutation rates have long been measured only as averages across many genomic positions (e.g.,  \citenum{Haldane1949,VogelMotulsky1997,Kondrashov2003,Segurel_etal2014,Rahbari_etal2016}). Recently, however, a method to measure the rates of individual mutations was developed \cite{Melamed_etal2022}. When this method was applied to a mutation that has served as a classic example of adaptation by random mutation and natural selection \cite{Allison1954,SforzaFeldman2003,FreemanHerron2007,HartlClark2007}---the human hemoglobin S (HbS) mutation, which protects against malaria while causing sickle-cell anemia in homozygotes \cite{Pauling_etal1949,Allison1954,Flint_etal1998,Kwiatkowski2005,Piel_etal2010}---it was found that this mutation originates de novo significantly more frequently in the gene and population case where it is adaptive: in sub-Saharan Africans, who have been exposed to intense malarial pressure for many generations, compared to northern Europeans, who have not, and in the hemoglobin subunit beta gene, where it provides protection against malaria, compared to the same mutation in the hemoglobin subunit delta gene, which does not \cite{Melamed_etal2022}. In other words, instead of originating with a probability unrelated to its usefulness and only then being favored by selection where it is adaptive, this mutation arises in the first place de novo more frequently where it is needed \cite{Melamed_etal2022}. In the second study of this kind, similar results were obtained for another mutation (the human \textit{APOL1} 1024 A$\rightarrow$G mutation), protective against the most common form of human African sleeping sickness \cite{Melamed_etal2025}. These results cannot be explained by previous models in evolutionary theory, because these models do not apply to the rates of individual DNA mutations.\footnote{As modifier theory shows, because mutation rates are low, for a mutation to increase the mutation rate and be favored for this effect by selection, it must increase the average mutation rate across many genomic positions, so that many mutations potentially induced by it would figure into its selective benefit \cite{HodgkinsonEyre-Walker2011,MartincorenaLuscombe2013,WalshLynch2018,Monroe_etal2022}. } Instead, the results support IBE, which holds that mutation origination at the mutation-specific level responds to information accumulated in the genome over generations \cite{Livnat2013,Livnat2017,LivnatMelamed2023}.

Modeling IBE realistically would be challenging. However, it is conceptually informative to examine estimation-of-distribution algorithms (EDAs) from an IBE perspective. In EDAs, a sample of possible solutions to a complex problem is generated; the better solutions in the sample are identified; a distribution from which these better solutions could have been drawn is estimated; this distribution is used to generate the next sample; and this procedure is repeated until a satisfactory solution is found (e.g., \citenum{Baluja1994,MuhlenbeinPaass1996,DeBonetIsbellViola1997,BalujaDavies1997,HarikLoboGoldberg1998,PelikanMuhlenbein1999,Harik1999,MuhlenbeinMahnig1999,EtxeberriaLarranaga1999,PelikanGoldbergCantuPaz1999,Probst_etal2017,DoerrDufay2022}). Here we show that multivariate EDAs---EDAs which allow for dependencies between variables\footnote{For our purposes, it is convenient to divide EDAs into two kinds: univariate and multivariate ones. The former do not allow for interactions between genes, whereas the latter do. Because bivariate EDAs allow for such interactions, we include them in the latter \cite{LehreNguyen2019,KrejcaWitt2019}, even though they have also been commonly defined as a separate category \cite{Pelikan_etal2002,LarranagaBielza2024}.}, i.e., for interactions between genes (e.g.,  \citenum{DeBonetIsbellViola1997,BalujaDavies1997,PelikanMuhlenbein1999,Harik1999,MuhlenbeinMahnig1999,EtxeberriaLarranaga1999,PelikanGoldbergCantuPaz1999,Probst_etal2017})---demonstrate in an abstract and indirect manner nonrandom, non-Lamarckian mutation. In particular, the model shows that nonrandom, non-Lamarckian mutations are natural: the key operator of the Restricted Boltzmann Machine which we use to generate the distribution \cite{Smolensky1986,Hinton2002}, which represents a kind of Hebbian learning \cite{Hebb1949,LowelSinger1992}, is analogous in the abstract to actual biological mutational mechanisms for which evidence exists \cite{Bolotin_etal2023,LivnatMelamed2023}.

Despite their long-standing contributions to evolutionary theory, analytical population genetic models of the classical form \cite{Fisher1930,Wright1931,Wright1932,Haldane1932} cannot answer the basic question of whether selection of random mutations, random genetic drift, recombination, etc., could account for the adaptive complexity of life given the amount of time that has been available for evolution, partly because they do not include a representation of the phenotype---of complex biological structure.\footnote{\textit{If} evolution \textit{is} based on random mutation and natural selection (rm/ns), then evidently time was sufficient for rm/ns to bring about life, but then rm/ns is taken as an assumption and is not mathematically justified.} Precisely in order to answer this question, Valiant introduced structure and means of analyzing complexity into evolutionary modeling by using the tools of theoretical computer science---specifically the ``probably approximately correct'' (PAC) learning framework \cite{Valiant2009}. In constructing his model, however, he followed the traditional dichotomy of random mutation vs. Lamarckism. Excluding the latter, he obtained that evolution is equivalent to a restricted form of Statistical Query (SQ) learning \cite{Kearns1998}---a process not obviously powerful enough to explain the evolution of life with---and argued that his model offered a unifying framework for evolution and cognition \cite{Valiant2009}. However, as we demonstrate here, it is when mutations are neither random nor Lamarckian that mutational mechanisms become analogous to operators of cognition \cite{Livnat2017,LivnatMelamed2023}, bringing evolution and cognition much closer together. 

Despite its abstract nature and differences from IBE, the model already captures indirectly several aspects of IBE, including that evolution is driven by the interaction of parsimony (simplification pressure) and fit \cite{Livnat2017,LivnatMelamed2023}; that the fundamental elements of evolution---mutation, recombination and selection---interact in a complementary fashion \cite{Livnat2013}; that random bits need not encode improvement directly but can enable generalization instead \cite{Vasylenko_etal2020}; and that improvement comes together with uniformity \cite{Livnat2013,Livnat2017}---the last of which offers a new angle on Darwin's puzzling observations that changed conditions lead to increased production of heritable variation \cite{Darwin1859}. Furthermore, while the bell-shaped distributions commonly observed in nature for continuous traits have been explained by the assumption of additive effects of multiple genes \cite{Fisher1918,FalconerMackay1996}, the results show that this shape can also be approximately obtained from a complex function where genes interact, and that its range of variation can be maintained for multiple generations under selection by a mechanism different than previously assumed \cite{Beatty2019}. Thus, the model points at a new direction for mathematical research on how to reconcile Darwinism and Mendelism in a manner that incorporates mutational mechanisms and reflects the central role of genetic interactions in biological evolution. 

Opening up new possibilities for thinking about how evolution happens is important for both evolution and computer science. For example, the mixability theory for the role of sexual recombination in evolution \cite{Livnat_etal2008,Livnat_etal2010,Livnat_etal2011,Vasylenko_etal2019,Vasylenko_etal2020} has served as a source of motivation for the development of ``dropout'' \cite{Srivastava_etal2014}---a regularization technique recognized as one of the breakthroughs that helped launch the deep-learning revolution (ref. \citenum{LeCun_etal2015}, p. 440). IBE shares multiple elements with mixability theory: the ideas that selection acts on complex wholes, that sexual recombination is not as a ``bonus’’ but a fundamental element of evolution, that the breaking down of genetic combinations is ``a feature, not a bug,'' and the importance of simplification and generalization in evolution  \cite{Livnat_etal2008,Livnat_etal2010,Livnat_etal2011,Vasylenko_etal2019,Vasylenko_etal2020}. However, because IBE involves also the role of mutation \cite{Livnat2013,Livnat2017,LivnatMelamed2023}, it is conceptually broader, and thus may serve as an additional source of motivation for interdisciplinary developments.

Our goal, therefore, is a pedagogical one. The results are not meant to provide a better analysis of EDAs in terms of convergence, runtime or comparisons to other algorithms and are not surprising from an EDA perspective (see, e.g., \citenum{LarranagaLozano2002,Pelikan_etal2002,HauschildPelikan2011,LehreNguyen2019,KrejcaWitt2019}). Our simulation does not model biological mechanisms of mutation realistically nor does it outperform preexisting evolutionary algorithms (e.g., \citenum{Shim_etal2013,Ceberio_etal2014,Liang_etal2015,Probst_etal2017}). Rather, our goal is to demonstrate that the traditional dichotomy of random mutation vs. Lamarckism is false, that the assumption of random mutation made throughout evolutionary theory is a contingent modeling choice, not a necessity, and that the concept of nonrandom, non-Lamarckian mutation is a  logically coherent alternative. This forms a new bridge between the evolutionary biology and evolutionary computation literatures, shows that concepts of IBE can be explored mathematically, and points at new directions for research in evolutionary theory, evolutionary computation and machine learning. Indeed, it highlights that a new way of thinking about how biological evolution happens is possible \cite{Livnat2013,Livnat2017,LivnatMelamed2023}.

\section{Conceptual Overview: Interaction-based Evolution}

Three basic points of IBE are as follows. \textit{a}) Sexual recombination, natural selection, and nonrandom, non-Lamarckian mutation interact in a complementary fashion \cite{Livnat2013}. \textit{b}) Evolution is driven by the interaction of parsimony and fit \cite{Livnat2017}. \textit{c}) Given nonrandom, non-Lamarckian mutation, the gradual process of the evolution of regulation leads to punctuated mutational change \cite{LivnatMelamed2023}. We briefly review these points and provide an empirical example in the following sections. 

\subsection{The interaction between the basic elements of evolution}

Three fundamental open problems in evolution have been as follows: What is the role of sex in evolution? How can complex adaptations evolve under natural selection? What is the fundamental nature of mutation? \cite{Livnat2013}. These problems have been considered largely separate from each other in standard  evolutionary theory \cite{Fisher1930,Wright1931,Bell1982,BartonCharlesworth1998,Merlin2010}. However, IBE holds that they reflect inseparable aspects of one and the same process \cite{Livnat2013}.

It is convenient to begin with the problem of sex. Sexual reproduction requires highly complex adaptations as well as time and energy investment across the scales of organization---from the complex molecular machinery of meiosis to flowering and animal behavior. Yet despite these costs, species in which individuals do not exchange genetic material in any form or manner appear not to last over evolutionary time \cite{Stebbins1950,Stebbins1957,Williams1966,VanValen1975}. These observations raise the following question: What makes sexual reproduction putatively necessary for long-term evolution?

A common intuition has been that the role of sex in evolution must be to generate a vast number of different combinations of genes. Because genetic variation is the ``fuel'' for natural selection, it is thought that this variation must speed up evolution. However, precisely how it may speed up evolution has been unclear \cite{EshelFeldman1970,Lewontin1971,BartonCharlesworth1998,West_etal1999}. In particular, sexual recombination cannot simply enable selection to favor the best particular genetic combinations among many, because it also breaks down the complex combinations that it generates \cite{EshelFeldman1970,Lewontin1971,BartonCharlesworth1998,LivnatPapadimitriou2016alg}.\footnote{For this reason, evolutionary theoreticians traditionally have been looking for other solutions to the problem of sex. Some of these proposed solutions do not consider genetic interactions to be important at all \cite{Fisher1930,Muller1932}, while others consider them important only in a narrow sense (e.g., arguing that rare genetic combinations escape common parasites and thus it is beneficial to keep regenerating them [\citenum{Hamilton1980, Jaenike1978}]).}

\begin{figure}[H]
\centering 
\includegraphics{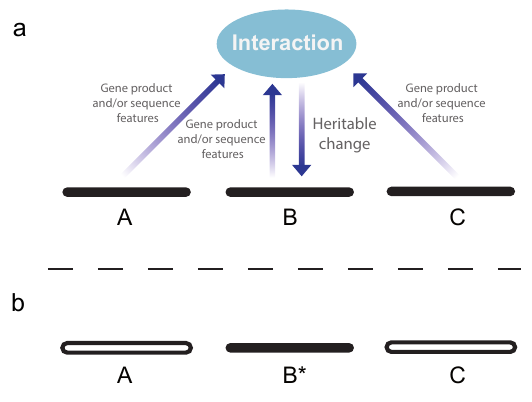}
\caption{The origination of heritable change is influenced by fan-in of heritable information, according to the theory of interaction-based evolution (image from ref. \citenum{LivnatMelamed2023}; see also ref. \citenum{Livnat2013}). \textit{a}) Gene products as well as genetic sequences arising from or residing at multiple recombining loci interact in influencing the origination of a heritable change, whether a DNA mutation or an epigenetic change. In this schematic figure, this coming together of information leads to a new mutation, B*. \textit{b}) As a result, even though the genes involved in the interaction may be separated by sexual recombination, the combination that they once were does have a heritable effect on future generations through the heritable change that was derived from it (B*). For simplicity, only three loci are presented, though in reality many loci can be involved in an interaction, directly or indirectly.}
\label{Fig1}
\end{figure}

According to IBE, this original intuition has been incomplete. IBE holds that heritable change integrates information \cite{Livnat2013,Livnat2017,LivnatMelamed2023,Melamed_etal2025}. Therefore, although complex genetic combinations are constantly broken down by sexual recombination, information from these combinations as complex wholes is transmitted to future generations through the heritable changes that are derived from them (Figure~\ref{Fig1}).

This recognition offers a new conceptual framework for the workings of evolution: sexual recombination generates a vast number of different transient combinations of genes, natural selection acts on these transient combinations as complex wholes, and information is transmitted from these combinations as complex wholes to future generations through the heritable changes that are derived from them. The probability of transmission accords precisely with the combinations' fitnesses \cite{Livnat2013}.

\begin{figure}[H]
\centering 
\includegraphics{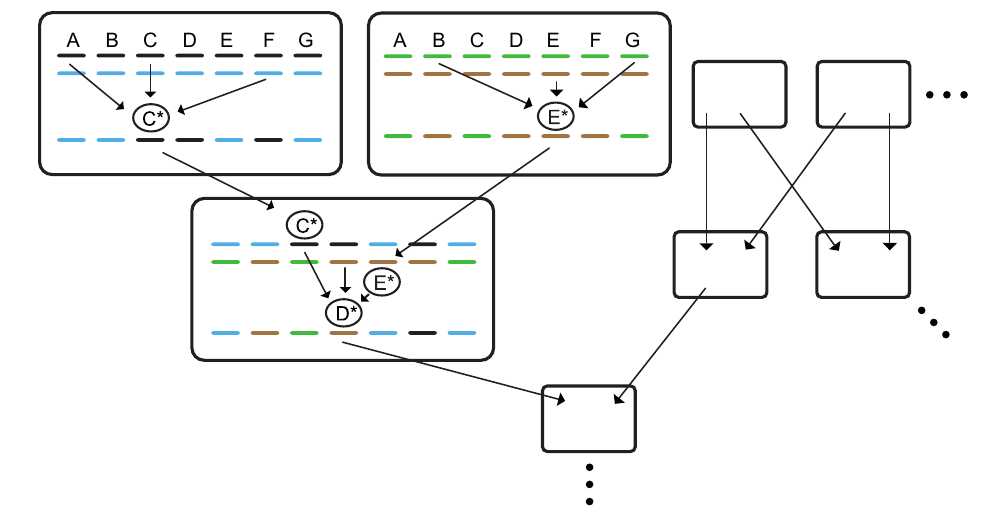}
\caption{The fan-in of information in the generation of heritable change enables a network of information flow across the genome and through the generations (image from ref. \citenum{Livnat2013}). Each box represents an individual. The large boxes represent two parents (top) and their offspring. The two dashed lines at the top of each large box represent the individual's diploid genome (each pair of dashes represents a diploid locus), and the dashed line at the bottom represents the haploid genome of a gamete that is transferred to the next generation. The figure demonstrates schematically that an output of a heritable change event at one generation (e.g., C*) can serve as an input into heritable change events at later generations (D*). This connectedness generates a network of information flow across the genome and through the generations \cite{Livnat2013}.}
\label{Fig2}
\end{figure}

Furthermore, the fact that the outcome of a heritable change event in one generation becomes a part of the genetic and epigenetic information that affects the probabilities of future heritable changes leads to a network of information flow and ``computation''\footnote{We place ``computation’’ in quotations because, despite the simple simulation to be presented here, biological evolution is not a single algorithm devised to solve a particular computational task.} across the genome and through the generations, where information flows from many genes into any one gene and from many ancestors into any one descendant (Figure~\ref{Fig2}).

\subsection{Evolution driven by parsimony and fit}

What can mutation ``compute’’ if no sensing of the immediate environment is allowed---if Lamarckism is excluded?

Under traditional theory, evolution is driven by a single  force---that of natural selection. In contrast, according to IBE, evolution is driven by the interaction of two forces---the external force of natural selection, and an internal force of natural simplification. The former represents differential survival and reproduction, and the latter represents nonrandom, non-Lamarckian heritable change \cite{Livnat2017,LivnatMelamed2023,Melamed_etal2025}. While the internal force enables simplification as well as duplication of heritable information \cite{LivnatMelamed2023}, the external force ensures that processes are only simplified to the extent that they keep functioning \cite{Livnat2017}. According to IBE, this interaction of parsimony and fit generates elements that not only perform better but also have the inherent capacity to come together with other such elements into emergent, useful interactions at the system level \cite{Livnat2017}. In other words, the interaction between parsimony and fit underlies the creativity of evolution \cite{Livnat2017,LivnatMelamed2023}.

This view addresses multiple fundamental questions. First, as to what mutations could do except for ``try changes at random,'' local simplification of internal information does not require ``knowledge'' of organismal fitness the way natural selection or Lamarckism do \cite{Livnat2017}: parsimony and fit interact in a complementary fashion---they do not overlap.

Second, it addresses from a new angle the fundamental question of how biological complexity arises. By arguing that simplification under performance pressure generates from preexisting interactions new elements that have the inherent capacity to engage in emergent interactions---unexpected, useful interactions at the system level---IBE suggests that simplification is key to complexity \cite{Livnat2017,Bolotin_etal2023,LivnatMelamed2023}.

Notably, ``co-option''---the phenomenon whereby a trait serving one role later in evolution becomes useful for another role---is ubiquitous  across the scales of biological organization  \cite{GouldVrba1982,Muller1990,MullerWagner1991,SchlichtingPigliucci1998,GraurLi2000,West-Eberhard2003,Hallgrimsson_etal2012}, and has been considered key to the emergence of complexity \cite{Muller1990,MullerWagner1991,SchlichtingPigliucci1998,West-Eberhard2003,MullerNewman2005,Moczek2008,Hallgrimsson_etal2012}. Like mutation, co-option has been thought to be accidental in standard evolutionary theory  \cite{Williams1966,GouldVrba1982,Gould2002}. However, according to IBE, both mutation and co-option have causes, and their causes are interrelated \cite{Livnat2017}: mutation embodies simplification, and simplification, together with selection, enables co-option \cite{Livnat2017}.

\subsection{The empirical nature of mutation and the connection between gradual evolution and punctuated mutational change}

IBE's third point connects the gradual process of the evolution of regulation with punctuated mutational change, thus addressing the HbS and other mutations' origination rates mentioned earlier \cite{Melamed_etal2022,LivnatMelamed2023,Bolotin_etal2023,Melamed_etal2025}. It holds that, starting from genetic variation initially co-opted from previous processes, under selection, evolution gradually focuses on the genomic regions and the particular regulatory interactions most relevant for the currently evolving adaptations---interactions from which large-effect mutations follow directly via mutational mechanisms \cite{Melamed_etal2022,LivnatMelamed2023,Melamed_etal2025}. Thus, adaptive mutations like the HbS or \textit{APOL1} 1024A$\rightarrow$G ones need not initiate a process of adaptation nor originate at random: they follow mechanistically from preexisting genetic interactions that have already evolved \cite{LivnatMelamed2023,Melamed_etal2025}. This means that the gradual and flexible process of adaptive evolution of regulation, which involves many changes of small effect, paves the way to  evolutionarily punctuated, structural mutational changes of large effect. This point connects between continuous and discrete evolutionary change \cite{LivnatMelamed2023,Melamed_etal2025}.

\subsection{An empirical example demonstrating principles of IBE}

Mutations come in various forms---not only the ``single-letter’’ changes called point mutations but also duplications, deletions, translocations, inversions and fusions of genetic sequences. \cite{GraurLi2000}. As for all mutation types, it has been thought that fusion mutations are random---that one day, a sequence is translocated by chance to another genomic locus, where by chance it may connect to a local sequence and form a new beneficial fusion gene. However, a recent study found that genes do not fuse at random \cite{Bolotin_etal2023}. Instead, genes that are used together repeatedly over the generations are the ones more likely to become fused together by mutational mechanisms based on their interaction \cite{Bolotin_etal2023}.

Mechanistically, it is known that when two remote genes work together, the DNA bends in 3D space, bringing them together to the same place at the same time in the nucleus with their chromatin open \cite{Jackson_etal1993,EdelmanFraser2012,Dixon_etal2012,Papantonis_etal2013,Soler_etal2017}. It is thus easy to see that reverse transcription of RNA, recombination-based mechanisms and other mechanisms may  specifically fuse such genes rather than others \cite{Livnat2017,LivnatPapadimitriou2016TREE}. Recent evidence supports this proposal \cite{Bolotin_etal2023}.

An analogy can now be made between the gene fusion mechanism and the workings of the brain, where, on the microscale, ``neurons that fire together wire together'' (Hebbian learning) \cite{Hebb1949,LowelSinger1992}, and, on the macroscale, pieces of information that are commonly used together become chunked or ``routinized'' into a single piece that is thereafter activated as one (this is one of the most basic principles of cognition and learning, called  ``chunking'') \cite{Lindley1966,TulvingCraik2005}.\footnote{For example, when we first learn to drive a manual car, switching gears requires conscious effort to coordinate different movements. However, in the course of learning, movements that repeatedly co-occur become chunked, allowing the gears to be shifted automatically and freeing the brain to learn higher-level tasks.} Thus, not only are gene fusion mutations nonrandom, their mechanism of origination is actually analogous to one of the most basic principles of cognition and learning by the brain.

Similarly to how the chunks generated by learning simplify processes and free up the brain to learn more complex patterns, gene fusion represents local regulatory simplification, and can facilitate evolution. Gene fusion does not lead to ever more simplicity and diminution of the genome; on the contrary. Because it is often preceded by or occurs simultaneously with duplication of the source copies, it often leads in the long term not from two genes to one but from two genes to three \cite{Livnat2017,Bolotin_etal2023}. This exemplifies the IBE principle that simplification is key to complexity: local regulatory simplification under selection, together with gene duplication, leads to a global increase in complexity  \cite{Livnat2017,LivnatMelamed2023}.

The gene fusion mutational mechanism demonstrates that the probability of a mutational event is mutation specific (each pair of genes has its own fusion rate) and is affected by complex information accumulated in the genome over generations, including all of the enhancers, promoters, transcription factors, epigenetic marks etc. across loci that regulate the two genes and determine the strength of the interaction between them. It follows that a large-effect, structural mutation can follow directly and mechanistically from a gradual process of evolution of regulation, such as the gradual tightening of the interaction between two genes \cite{LivnatMelamed2023,Bolotin_etal2023,Melamed_etal2025}.

It has been proposed that, not only for gene fusion mutations, but across mutation types, mutation rates are mutation-specific and dependent on pre-evolved information in a biologically meaningful manner, and that mutational mechanisms put together different pieces of heritable information and simplify evolved regulatory phenomena \cite{LivnatMelamed2023}.

\section{General overview of the simulation} 

Modeling IBE realistically is a challenging task for future research. Here, we point out that the multivariate EDA framework captures nonrandom, non-Lamarckian mutation and certain other aspects of IBE in an abstract and indirect manner.

In our EDA model, a Restricted Boltzmann Machine (RBM) \cite{Smolensky1986,Hinton2002} brings together information from the genomes of all individuals at each generation, representing at an abstract level the interaction between sexual recombination and nonrandom, non-Lamarckian mutation. Obviously, this use of the RBM is unrealistic. IBE does not hold that information comes together from all individuals at once per generation, nor that mutation and recombination can be literally represented by an RBM. We use the RBM as a generic representation of statistical learning, not as a claim about specific biological mechanisms, and its simultaneous pooling of information is a rudimentary stand-in for more natural forms of recombination and mutation that may be modelled in the future.

The simulation comprises an adaptational problem to be approximately solved by the evolution of a population of constant size of $N$ genomes per generation. Each genome is a possible solution to the adaptational problem and is assigned a fitness score based on its quality as such a solution. At each generation, threshold selection is applied, such that a certain top percentage (e.g., top 50\%) of the population survive. The surviving individuals reproduce with change to provide the next generation of $N$ genomes. The simulation starts with randomly produced genomes, and, over the generations, the average fitness in the population increases, as expected from standard models.

\begin{figure}
\centering
\subfloat{\includegraphics[trim=0cm 4cm 0cm 3.8cm, clip=true, scale=0.6]{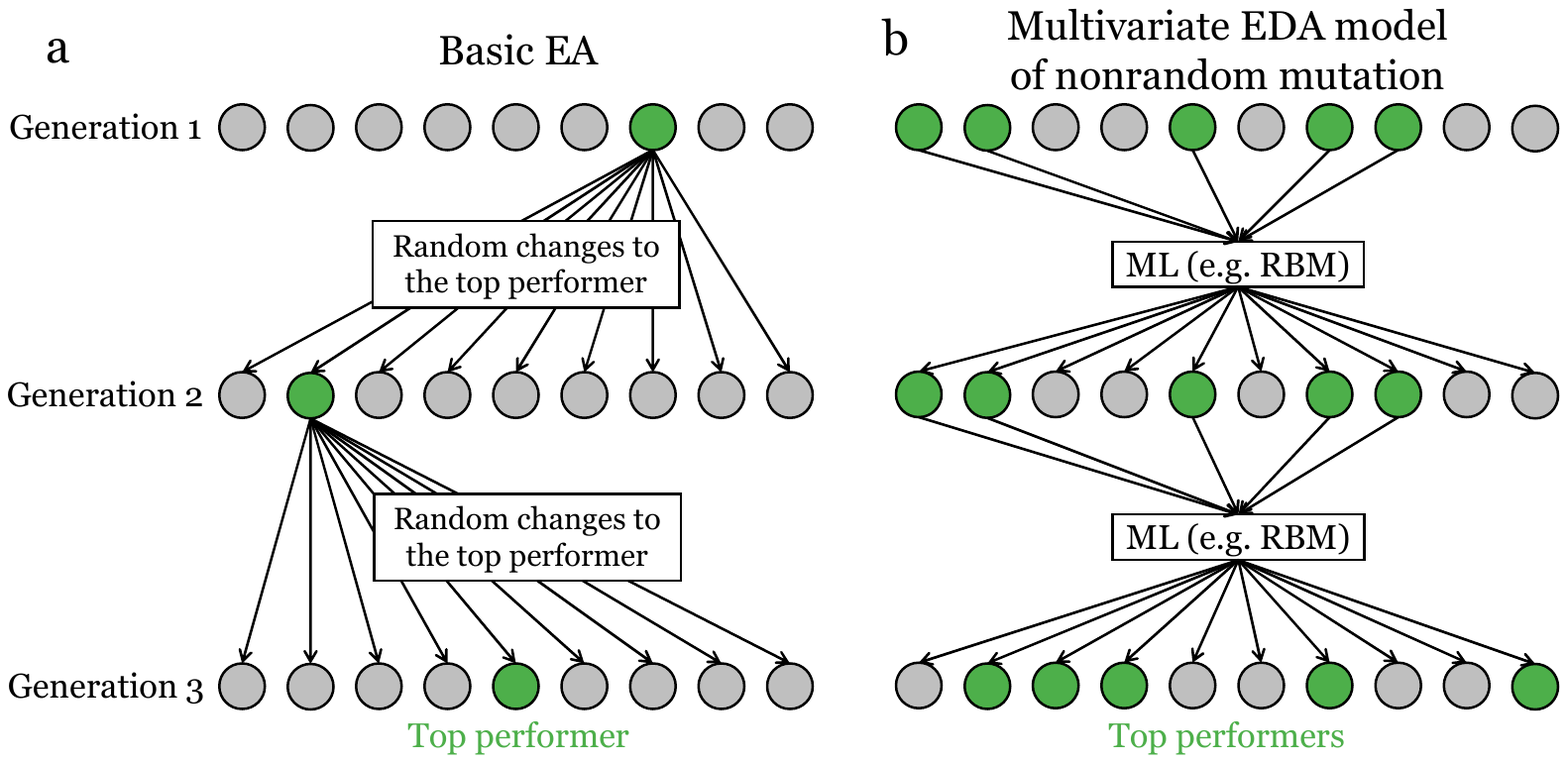}}\\
\subfloat{\includegraphics[trim=0.4cm 9.5cm 9.8cm 3.5cm, clip=true, scale=0.6]{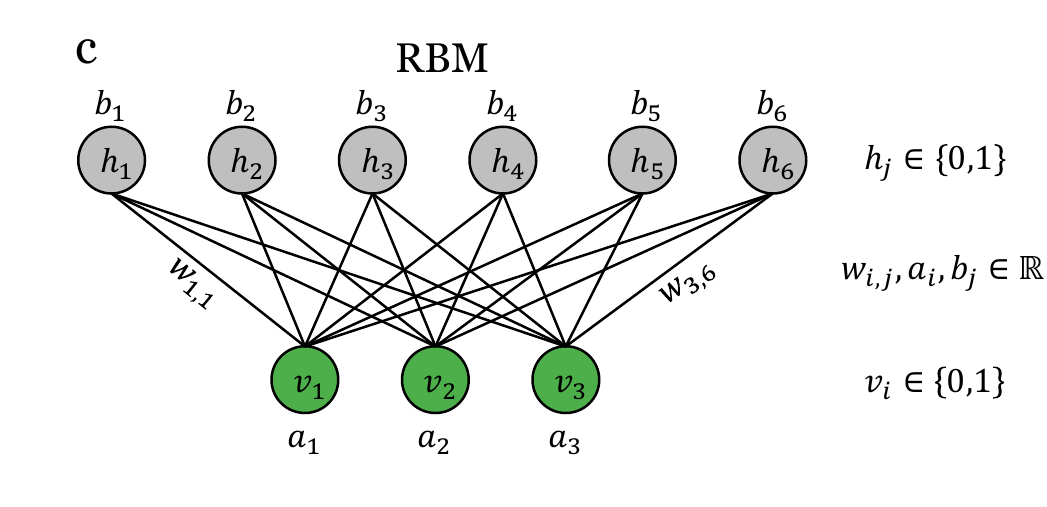}}
\caption{A comparison between basic evolutionary algorithms (EAs) and the RBM-based EDA model described here. \textit{a}) A schematic diagram of a basic evolutionary algorithm. At each generation, the top performer is selected and produces the next generation via self-replication and random changes representing mutations. \textit{b}) A schematic diagram of the RBM-based EDA model. At each generation, a certain percentage of top performers is selected. Indirectly representing sexual recombination and nonrandom, non-Lamarckian mutation, an RBM is trained on the genomes of these top performers and, following training, is used to produce the next generation. \textit{c}) An example of a Restricted Boltzmann Machine (RBM) with three visible units $v_i$, six hidden units $h_j$, $18$ weights $w_{i,j}$ between them, and nine biases in total for visible ($a_i$) and hidden ($b_j$) units. In the simulations to follow, the number of visible units equals the genome length, which can be much larger than three. During training, the genomes of the surviving individuals are fed one by one into the visible units, and the weights and biases are adjusted accordingly (see Section 4 for details).}
\label{Fig3}
\end{figure}

The difference from basic genetic algorithms (GAs) and standard evolutionary models \cite{Fisher1930,Wright1931,Wright1932,Haldane1932,Holland1975,DeJong1975,Goldberg1989,Droste_etal2002} is in how individuals reproduce with change. At each generation, all surviving genomes are used to train a machine learning (ML) model, allowing it to learn the underlying distribution of such genomes. Next, $N$ new individuals are sampled from the model to form the next generation. The process repeats itself for a predetermined number of generations, and the fitnesses of the individuals are recorded over the generations (Figure~\ref{Fig3}). As the ML component, we used here an RBM \cite{TangShimTanChia2010,ShimTanChia2010,ShimTan2012,ShimTanChiaMamun2013,Shim_etal2013,Probst_etal2017,BaoSunGongZhang2022},though other options are possible (e.g., \citenum{PelikanGoldbergCantuPaz1999,Probst2016,ProbstRothlauf2020,JeongLeeLeeAhn2022}).

The use of this multivariate EDA framework reflects a key difference between IBE and previous evolutionary thinking. In the latter, there is no internal fan-in of information---no coming together of separate pieces of information found at different recombining loci in the genome and in different individuals by mutation. Any computational integration of information can only occur through natural selection as it evaluates the phenotype. In contrast, at the core of our simulation, there \textit{is} such internal fan-in, represented by the ML component---in principle, albeit not in realistic detail, in accord with IBE \cite{Livnat2013,Livnat2017}.

\subsection{How the simulation captures IBE principles}

The model demonstrates heritable change that is neither random nor Lamarckian. It is not random, because new genomes are produced as a result of computation done on previous genomes. It is not Lamarckian, because the RBM cannot sense the environment or respond to it. Indeed, the RBM is not exposed to the selection function. It can only operate on internal information in the genomes of individuals.

Despite this abstract manner of modeling, we can already see that the heritable change represented by the RBM is actually natural in a critical sense. This can be seen by the analogy between the internal workings of the RBM and the biological gene fusion mutational mechanism mentioned earlier. In Boltzmann Machines, units that are used together are more likely to have the connection between them strengthened. In biological evolution, genes that are used together are more likely to become fused together.\footnote{While in the RBM there are no direct connections between visible units, indirect connections via hidden units can be strengthened. The difference between fusion of units and the strengthening of connections between them is also mild for our current pedagogical purpose \cite{Livnat2017}: as one example, it has been argued that genes that are used together are not only more likely to become fused together but also to be translocated closer to each other, long before they become fused \cite{Bolotin_etal2023}, indicating a more gradual evolutionary process.} Though more detail at the molecular level may become known in the future, the analogy can already be made at an abstract level and demonstrates that the mechanisms of nonrandom, non-Lamarckian heritable change are natural: they are local, with fan-in, and supported by empirical evidence \cite{LivnatMelamed2023}.

In the sense that recombination brings information from separate surviving genomes into one, the EDA framework captures the idea of sexual recombination at a high level (Figure~\ref{Fig3}). Indeed, EDAs have been classified under ``recombination-based search methods'' \cite{Rothlauf2011}. In addition, in the multivariate case, the EDA framework captures the IBE principle of an interaction between sexual recombination and nonrandom, non-Lamarckian mutation, in that the internal fan-in through the RBM enables the coming together of information from different genes and different ancestors into any one gene and any one descendant (Figure~\ref{Fig2}).

\section{Details of the simulation}

\subsection{Boltzmann Machines, Hebbian learning, and the used-together-fused-together principle}
\label{BM}

A Boltzmann Machine (BM) \cite{Ackley_etal1985,HintonSejnowski1986} is a network that can be used, among other things, to form an implicit estimate of the probability distribution underlying a sample and then generate more data points from this estimated distribution. At each point in time, each of the units in the network can be either in an on (1) or in an off (0) state, and these states are updated asynchronously based on the states of other nodes in the network at the last time point. Let 
\begin{equation}
\Delta E_k=\sum_i w_{ki}s_i - b_k,
\label{DelE}
\end{equation}
where $w_{ki}$ is the weight of the bidirectional connection between nodes $k$ and $i$ ($w_{ki}=w_{ik}$), $s_i$ is the state of node $i$, and $b_k$ is a bias term. In the probabilistic setting \cite{Ackley_etal1985}, at each time point, a node $k$ is chosen at random with replacement, and its state $s_k$ is updated to 1 with probability 
\begin{equation}
p_k=\frac{1}{1+e^{-\Delta E_k/T}},
\end{equation}
where $T$ is a ``temperature'' parameter, and otherwise is set to 0. Adding a unit that is constitutively on and whose connection to node $k$ has the value $-b_k$ simplifies equation \ref{DelE} to 
\begin{equation}
\Delta E_k=\sum_i w_{ki}s_i,
\end{equation}
showing that the bias can be updated like any other connection weight by the training algorithm, as discussed below.

The network consists of visible units, which serve as the input/output interface, and hidden units. During training, the different data points are fed into the visible units, each data point being a binary vector $X_j\in\{0,1\}^n$, and, after training, the network is allowed to run freely and generate data points on the visible units in accord with the generative model learned.

During training, the network is run in two different modes: In the clamped mode, the visible units are temporarily fixed to the values of a point in accord with its probability, and the network states are allowed to reach an equilibrium in accord with an annealing schedule,  where the temperature is gradually reduced. In the free-running mode, random values are fed into the visible units, and these values are allowed to change while the network is allowed to reach an equilibrium. For each pair of units $\{i,j\}$, the probabilities of both units being on at the same time across equilibria in  the clamped mode, denoted $p_{ij}$, and in the free-running mode, denoted  $p'_{ij}$, are calculated, and the weight of the connection between the units is adjusted according to the following rule:
\begin{equation}
\Delta w_{ij}=\frac{\eta}{T}(p_{ij}-p'_{ij}),
\label{wij}
\end{equation}
where $\eta$ is the learning rate parameter. In other words: if units tend to be on together at the same time at equilibrium, the regulatory connection between them is strengthened, making them activate each other more strongly in the future. This represents Hebbian learning---units that ``fire together wire together'' \cite{Hebb1949,LowelSinger1992}---and is conceptually related to the gene fusion mechanism described earlier \cite{Bolotin_etal2023}.

This operation represents local change with fan-in. Though it depends only on the two nodes it connects, this suffices to enable a form of simplification: the complex causes of the correlation between two nodes are gradually replaced with a local, simpler connection. Finally, note that the connection is strengthened more the farther the correlation between $i$ and $j$ in the clamped mode is from that correlation in the free-running mode.

Weights are adjusted in this manner until the discrepancy between the empirical distribution of examples and the generated distribution of examples, $$\sum_\alpha \frac{P(V_\alpha)\log P(V_\alpha)}{P'(V_\alpha)},$$ is sufficiently small, where $P(V_\alpha)$ and $P'(V_\alpha)$ are the probabilities of the visible nodes being at state $V_\alpha$ in the clamped and free-running states respectively (for detail, see refs.   \citenum{Ackley_etal1985,Yegnanarayana2003}).

A Restricted Boltzmann Machine (RBM) \cite{Smolensky1986,Hinton2002} is a simple form of BM that can be trained and executed much faster. In the RBM, edges between two hidden units as well as edges between two visible units are excluded, resulting in a single layer of visible nodes and a single layer of hidden nodes. Thus, fewer calculations are needed, and these calculations can be done in parallel. Furthermore, convergence and annealing are not used; instead, the weights are updated for a predetermined number of iterations, which is a model parameter (see below).

A basic training approach for the RBM, called Contrastive Divergence (CD-k) \cite{Hinton2002}, operates as follows. In the clamped mode, the states of hidden units are calculated only once, and the states of all units are noted. In the free-running mode, one may begin with random values or zeros in one of the layers, then calculate the states of the other layer, and next calculate the new states of the first layer. After a small number $k$ of such cycles, called ``Gibbs steps,'' the states of all units are noted. Based on the states noted in the two modes for all input vectors used, $p_{ij}$ and $p'_{ij}$ are calculated, and the weights of the connections between the layers are updated as before (equation \ref{wij}). Each computational cycle resulting in weights update is called ``an iteration.''

Here we use the RBM model with the scikit-learn implementation \cite{Pedregosa_etal2011}, which employs the Persistent Contrastive Divergence (PCD) method \cite{Tieleman2008}. PCD builds upon the CD-k approach, where, moving from one iteration to the next, the states in the free-running mode are not initialized, but instead the calculation continues from the states of the previous iteration \cite{Tieleman2008,TielemanHinton2009}. It has been suggested that PCD is more efficient than CD-k \cite{Tieleman2008}. Notice that, while the training of BMs normally begins with zero weights \cite{Ackley_etal1985} or with random weights in the range $[-1, 1]$ (ref. \citenum{Yegnanarayana2003}, p.~191), in the scikit-learn implementation, the weights and biases are initially sampled from the normal distribution $\mathcal{N}(0,\;0.01^2)$ and are then updated as described above.

As recommended \cite{Hinton2010}, we divide the input vectors into sets of equal size, called mini-batches. For each batch, we run the number of iterations specified and then continue to update the weights with the next batch. The number of hidden units, the learning rate, the number of iterations and the batch size are model parameters.

Following training, we use the free-running state of the RBM to generate vectors that accord with the rules discovered. To generate each new vector, we place a random binary value at each of the visible units ($0$ or $1$ with equal probability), run the RBM at a free state for a specified number of iterations and obtain the new vector at the visible nodes.

\subsection{Adaptational challenge and fitness values}

To generate complex adaptive landscapes, we use the well-known MAX-SAT problem \cite{Papadimitriou1994}, for which the performances of different EDAs have been compared before \cite{Probst_etal2017}. In MAX-SAT, a Boolean formula in conjunctive normal form (CNF) is given, e.g., $\phi=(x_1 \vee x_2) \wedge (\overline{x_1} \vee x_2 \vee x_3) \wedge (\overline{x_1} \vee \overline{x_2} \vee x_3)\wedge (\overline{x_2} \vee \overline{x_3})$, and one must find an assignment to the $x_i$ that maximizes the number of clauses that are satisfied. For example, the assignment $x_1=1$, $x_2=0$ and $x_3=1$ satisfies all 4 clauses in $\phi$. When all clauses in formula $\Phi$ are of size $k$, we call $\Phi$ a MAX-$k$-SAT instance.

Consider a haploid genome of $n$ loci, carrying at each locus one of two alleles, $0$ or $1$. Let fitness be a binary function $f_\Phi :\{0,1\}^n \rightarrow [0,1]$, which, for genome $i$, $X_i \in \{0,1\}^n$, provides the fraction of clauses in formula $\Phi$ that $X_i$ satisfies. In other words, each individual genome offers a solution to the MAX-SAT problem. While MAX-SAT is NP-hard, we will examine approximate solutions that are discovered in the course of two evolutionary processes: a random mutation and natural selection (rm/ns) process (RM), and an RBM-based, IBE-like process as described.

\subsection{The core of the simulation}

Consider a population of genomes $X_i$, $i \in \{1,...,N\}$, where $N$ is even for convenience. At generation $0$, all genomes are randomized such that each bit of each genome is either $0$ or $1$ with equal probability.

At each generation, the fitnesses of all individuals are calculated. In the IBE-like case, a top percentage (e.g., $50\%$) of the population survives. An RBM with $n$ visible units is then trained on the surviving individuals as explained above, and is then run in a free state to generate all $N$ individuals of the next generation.

In the rm/ns case, a certain percentage or number of top performing individuals survive. To generate each of the $N$ individuals of the next generation, a surviving parent is chosen at random with replacement, a copy of its genome is made, and, to simulate mutation, a specified number of bits in this copy are chosen at random with uniform probability and flipped.

The population mean fitness, $\bar{w}$, is calculated at each generation before selection, to avoid bias due to the percentage of surviving individuals parameter.

Note that the population size $N$, genome size $n$ and fitness function $f_\Phi$ parameters are shared between the models. The number of hidden units $h$, the number of iterations $T$, the batch size $B$, the learning rate $\eta$ and the RBM number or percentage of surviving individuals $S_{RBM}$ are specific to the RBM model, while the number of bits mutated per individual per generation $\mu$ and the RM number or percentage of surviving individuals $S_{RM}$ are specific to the RM model.

In addition to the RM and RBM models, a third, basic model called Best Random Guess (BRG) is run. This model generates $N$ random genomes of size $n$ at each generation and stores in memory the genome with the highest fitness value observed up to that generation. It demonstrates the solution quality that can be reached by repeated pure guessing.

\subsection{Problems used}
\label{subsection_problems}

\noindent \textit{a}) Uniform Random $k$-SAT Instances: The random k-SAT generator \cite{Balint_etal2012} was used to create uniform random MAX-k-SAT instances with $k = 3$, different numbers of variables $n$ ranging from $75$ to $2400$ and a clause-to-variable ratio of $4.267$. For each $n$, the tool randomly selects three variables from the set of $n$ variables and their negations, repeating this process $n \cdot 4.267$ times, rounded up to the nearest integer, while ensuring that, within each clause, all variables are distinct, and that all clauses are distinct from each other.

\vspace{3mm}

\noindent \textit{b}) Even Boolean Parity Functions: As discussed below, to examine whether the RBM uses information on interactions between genes---whether it determines the genetic content at one genomic locus based on another---we used even-parity Boolean functions. An even-parity function outputs $1$ when the number of ones among its Boolean variables is even, and $0$ otherwise. These functions have known satisfying assignments which are distinct from each other. For example, the two-variable even-parity problem has two satisfying assignments: $(0,0)$ and $(1,1)$. We used even-parity Boolean functions with $2$--$5$ variables.

\subsection{Parameter choice}

To enable a meaningful comparison of the performances of the two evolutionary models in the adaptational challenge, for each setting of the shared parameters, we run each model with its best tested model-specific parameters. Because the underlying question is whether nonrandom, non-Lamarckian mutational mechanisms are possible in principle, the RBM is provided as much computational power as needed within the limit of the computational resources available for the study.

Allowing each model its best parameters is conservative for our purposes. In standard evolutionary theory, the rate of random mutation does not adjust itself to any particular combination of population size, genome size and adaptational challenge. Therefore, the optimal choice of the random mutation rate per shared parameter setting makes the RM model far more powerful than expected from standard theory. In contrast, according to IBE, mutational and recombinational mechanisms evolve in an ongoing manner and, although it is beyond the scope of the present work to model their own evolution, they are expected to fit the evolution of current adaptations. Given that our simulation allows each model to run with its best parameters, the fact that the evolution of the mutation rate is expected to be severely restricted from the perspective of traditional theory, whereas mechanisms of mutation and recombination are expected to continually evolve from the perspective of IBE, makes this comparison conservative in principle.

To ensure that it is conservative in practice, however, it is also crucial to test that the parameter range scanned in order to optimize the RM model is a reasonable one. For this purpose, we have investigated trends in the RM model for the six MAX-$3$-SAT  examples described above using various population sizes, percentages of surviving individuals and numbers of mutations per individual per generation:
\begin{enumerate}
	\item Population size, $N$: $1$, $10$, $100$, $1000$, $10000$, $100000$.
	\item Percentage of surviving individuals, $S_{RM}$: $1\%$, $5\%$, $10\%$, $50\%$ and $1$ individual.
	\item Number of mutations per generation, $\mu$: $1-5$.
\end{enumerate}

All six MAX-$3$-SAT cases exhibit similar trends. As an example, Table~\ref{Trends_unif-k3-1200_SR_gen500_Av} presents the average fitness values of populations at generation $500$ for a MAX-$3$-SAT problem with $1200$ variables and all possible combinations of the parameters described above. We can see that, first, for the given population sizes, which are the ones used in all of the simulations to follow, the optimal number of mutations per individual per generation varied from $1$ to $4$, suggesting that testing numbers of mutation larger than the ones tested in the simulations to follow is unnecessary. Second, the best results are typically observed when $1$ individual, $1\%$ or $5\%$ of the population survives. More rarely, depending on the particular problem, for larger population sizes, a survival rate of $10\%$ can outperform scenarios where only the best individual survives. This suggests that testing percentages of surviving individuals that are higher than tested in the simulations to follow is unnecessary.

\begin{table}[H]
\caption{Population mean fitness values at generation $500$ in the RM model for a MAX-3-SAT example with $n=1200$ variables. Some cells are blank because, for the corresponding population size and percentage of surviving individuals, either one or no individuals survive (if one individual survives, the results are presented in the top block of the table, and if no individual survives, the simulation cannot be run). In the first two rows, the results are better with $\mu=2$ than with $\mu=1$ for larger population sizes. The reason is that when only one individual survives, only $n$ different individuals can be produced by flipping only a single bit in the genome per offspring. Therefore, increasing $N$ beyond $n$ will not improve the results---it will merely generate duplicate individuals---unless $\mu$ is increased as well. This kind of effect applies only when the number of surviving individuals is small relative to the population size, so that the initial variation from which the next generation is produced is small. At the same time, increasing the percentage of surviving individuals is typically detrimental for the RM model.}
\begin{center}
\label{Trends_unif-k3-1200_SR_gen500_Av}
\begin{tabular}{||c|c||c|c|c|c|c||}
\hline
\multicolumn{2}{||c||}{population size, $N$ $\rightarrow$} & $2$ & $10$ & $100$ & $1000$ & $10000$ \\
\cline{1-2}
 $S_{RM}$ & $\mu$ &&&&& \\
\hline\hline
 & 1 & 0.919 & 0.974 & 0.986 & 0.986 & 0.985 \\
\cline{ 2 - 7 }
 & 2 & 0.917 & 0.966 & 0.986 & 0.99 & 0.989 \\
\cline{ 2 - 7 }
1 & 3 & 0.925 & 0.957 & 0.981 & 0.989 & 0.991 \\
\cline{ 2 - 7 }
individual & 4 & 0.925 & 0.947 & 0.976 & 0.987 & 0.991 \\
\cline{ 2 - 7 }
 & 5 & 0.927 & 0.948 & 0.973 & 0.984 & 0.989 \\
\hline
 & 1 & & & & 0.989 & 0.99 \\
\cline{ 2 - 7 }
 & 2 & & & & 0.99 & 0.993 \\
\cline{ 2 - 7 }
1\% & 3 & & & & 0.987 & 0.991 \\
\cline{ 2 - 7 }
 & 4 & & & & 0.982 & 0.986 \\
\cline{ 2 - 7 }
 & 5 & & & & 0.979 & 0.982 \\
\hline
 & 1 & & & 0.987 & 0.99 & 0.988 \\
\cline{ 2 - 7 }
 & 2 & & & 0.982 & 0.987 & 0.987 \\
\cline{ 2 - 7 }
5\% & 3 & & & 0.979 & 0.98 & 0.982 \\
\cline{ 2 - 7 }
 & 4 & & & 0.968 & 0.972 & 0.977 \\
\cline{ 2 - 7 }
 & 5 & & & 0.965 & 0.97 & 0.971 \\
\hline
 & 1 & & & 0.981 & 0.988 & 0.988 \\
\cline{ 2 - 7 }
 & 2 & & & 0.976 & 0.98 & 0.983 \\
\cline{ 2 - 7 }
10\% & 3 & & & 0.968 & 0.973 & 0.977 \\
\cline{ 2 - 7 }
 & 4 & & & 0.963 & 0.967 & 0.97 \\
\cline{ 2 - 7 }
 & 5 & & & 0.957 & 0.963 & 0.963 \\
\hline
 & 1 & & 0.939 & 0.957 & 0.96 & 0.962 \\
\cline{ 2 - 7 }
 & 2 & & 0.935 & 0.945 & 0.95 & 0.951 \\
\cline{ 2 - 7 }
50\% & 3 & & 0.929 & 0.936 & 0.943 & 0.943 \\
\cline{ 2 - 7 }
 & 4 & & 0.923 & 0.933 & 0.936 & 0.937 \\
\cline{ 2 - 7 }
 & 5 & & 0.92 & 0.927 & 0.93 & 0.931 \\
\hline\hline
\end{tabular}
\end{center}
\end{table}

\section{Results}
 
\subsection{Adaptive evolution with nonrandom, non-Lamarckian mutation}

\begin{figure}[H]
\centering 
\subfloat{\includegraphics[trim=0cm 6.8cm 1.4cm 8.7cm, clip=true, scale=0.83]{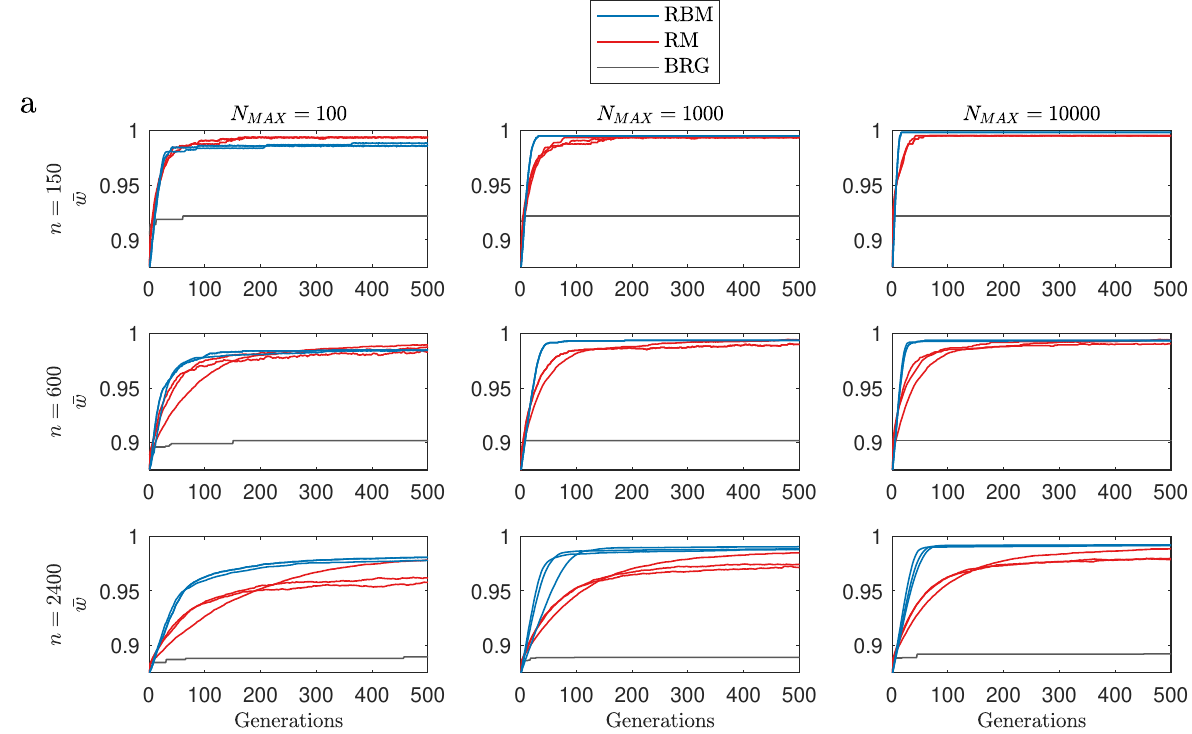}} \\
\subfloat{\includegraphics[trim=0cm 6.8cm 1.4cm 10.3cm, clip=true, scale=0.83]{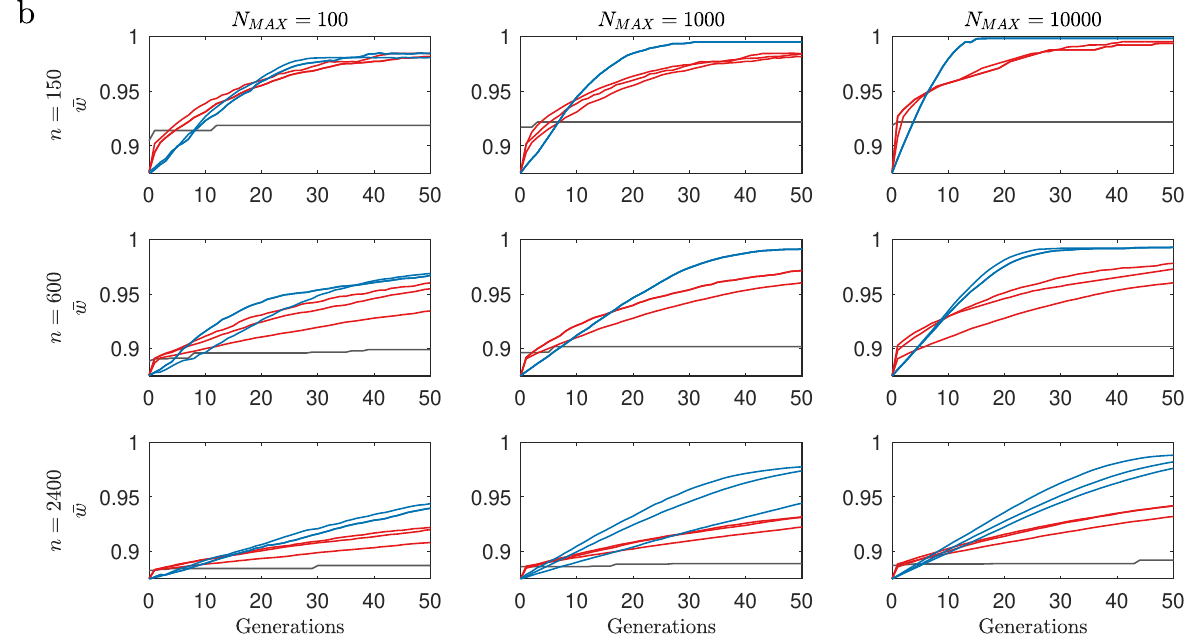}}
\caption{\textit{a}) $\bar{w}$ in the RBM (blue), RM (red) and BRG (grey) models for three MAX-3-SAT instances of  $150$, $600$ and $2400$ variables (rows) and three maximum population sizes $N_{max}$ (columns). Generations are on the $x$-axis. The difference between the RBM and RM mean fitnesses increases with the population size as well as with the problem size. The BRG curve, which provides a baseline for comparison, starts higher than the other curves because it represents not the mean fitness but the single highest fitness obtained up to the given generation by pure guessing. \textit{b}) Zooming in on the first $50$ generations. The RM model performs better during the first few generations, while the RBM model exhibits a rapid transition during the transient.}
\label{Fig4}
\end{figure}

We examined various MAX-SAT problem instances within the limits of the computational resources available, with different clause sizes, both fixed and variable, and different total numbers of variables. As a typical example, Figure~\ref{Fig4} shows three MAX-3-SAT instances with $150$, $600$ and $2400$ variables. To construct this figure, for each model, we first evaluated all possible combinations of the parameters shown in Table~\ref{Table_parameters} with the following exceptions. For the RM case, combinations with $N=10$ and $S_{RM}< 10\%$ were excluded because they allow less than 1 individual to survive. For the RBM case, combinations with less than 50 surviving individuals were excluded, because the RBM cannot be trained on insufficiently large datasets. For each model, we then found the parameter combinations that provided the highest population mean fitness at generations 50, 100 and 500 (up to three different parameter combinations---one for each of the three sampled generations) and, in the case of ties, picked one of the tied combinations at random. We then presented up to three curves for each model, one for each of the three best parameter combinations. To avoid the RM model being encumbered by a population too large to cross fitness valleys, for a given value $N_{MAX}$, the population size $N$ is itself an optimized parameter that is allowed to take any of the $N$ values in the table that are smaller or equal to $N_{MAX}$.

\begin{table}[H]
\caption{Parameters used in the RM and RBM models.}
\begin{center}
\label{Table_parameters}
\begin{tabular}{|c|c|c|}
\hline
Parameter & RBM & RM \\
\hline\hline
Problem & 3-SAT & 3-SAT \\
\hline
Population size, $N$ & $100, 1000, 10000$ & $2, 10, 100, 1000, 10000$ \\
\hline
Percentage of surviving individuals, & $5\%$, $10\%$, $50\%$ & $1$ individual, $1\%$, \\
$S_{RBM}, S_{RM}$ && $5\%$, $10\%$, $50\%$ \\
\hline
Number of hidden nodes, $H$ & $1\times, 2\times$ problem size & N/A \\
\hline
Number of iterations, $T$ & $20, 100$ & N/A \\
\hline
Learning rate, $\eta$ & $0.00001$, $0.0001$, & N/A \\
 & $0.001$, $0.01$, $0.05$ & \\
\hline
Batch size, $B$ & $10$, $100$, $1000$ & N/A \\
 & $B < N$ & \\
\hline
Number of mutations && \\
per individual per generation, $\mu$ & N/A & $1,2,3,4,5$ \\
\hline\hline
\end{tabular}
\end{center}
\end{table}

As Figure~\ref{Fig4}a shows, adaptive evolution proceeds effectively in the RBM-based model, namely with heritable change that is neither random nor Lamarckian. In fact, $\bar{w}$ becomes higher in the RBM model than in the RM model during the transient---i.e., following several (5--7) generations and before plateauing (at about generation 400--500). This RBM advantage increases with the problem size and/or the population size, suggesting that the RBM-based model solves the optimization problem more effectively than the RM model under more realistic conditions.

While the RM model performs better in the first few generations, the RBM model exhibits a rapid transition during the transient, with an inflection point in low population sizes and small problem instances.

\subsection{Genome completion test}

To ascertain that genes interact in the generation of heritable information, we took two approaches. The first involves using the RBM to complete partially known genomes. In this approach, we clamp values (henceforth ``known gene values'') to some of the visible units and run the RBM with these values fixed for $T$ iterations. This procedure allows the remaining visible unit activations (henceforth ``inferred gene values'') to be inferred from the fixed ones based on the patterns learned from surviving genomes. We perform this procedure at each generation separately following the original simulation described in Section 4. Each set of known gene values comes from one individual, and all sets involve the same loci.  After generating 100 completed genomes, we generate the set of all possible shuffled genomes, each of which includes one known and one inferred set of genes, regardless of the original linkage between them, resulting in $\frac{100 \cdot 99}{2} = 4950$ shuffled genomes in total. Finally, we compare the mean fitness of the inferred genomes to the mean fitness of the shuffled genomes. If the former is consistently larger, it would demonstrate that the inferred gene values match the known gene values---i.e., that information at one locus is generated based on information at other loci.

\begin{figure}
\centering 
\includegraphics[trim=2.4cm 6.0cm 2.6cm 8.1cm, clip=true, scale=0.9]{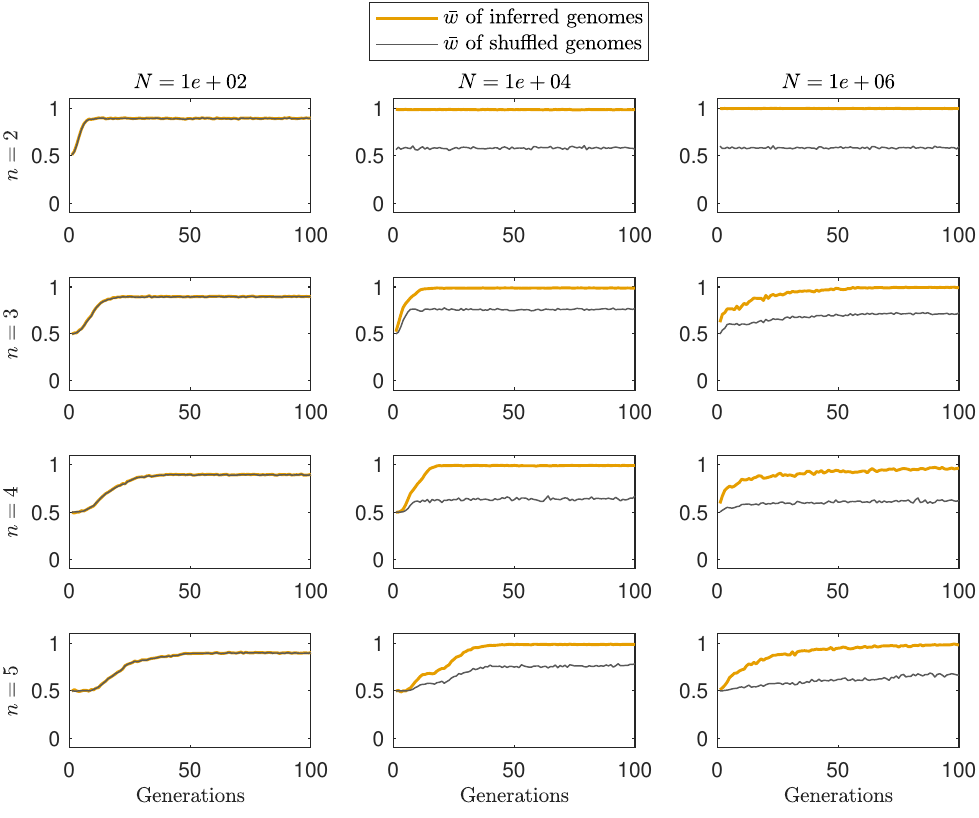}
\caption{$\bar{w}$ over the generations in the genome completion test for even-parity Boolean functions with $2$, $3$, $4$ and $5$ variables, averaged over 100 independent runs. The known gene values are sampled from the population and the remaining gene value is completed by the RBM. The rows show results for functions of different numbers of variables, from $n = 2$ to $n = 5$. The columns show results for different population sizes, from $N = 10^2$ to $N = 10^6$. The other parameters are $B = 10$, $T = 20$, $H = n$, $\eta = 0.1$, $S_{RBM} = 50\%$; the number of genomes to complete is $100$, and the length of the pattern to be completed is one position shorter than the genome length. The average fitness of the inferred genomes is shown by an orange line, and the average fitness of shuffled genomes is shown by a grey line. In the left column, the population size is generally too small to learn several different solutions, and therefore the shuffled genomes are identical to the inferred genomes. This effect disappears for larger population sizes, showing a gap in $\bar{w}$ between the inferred and shuffled genomes, which implies that information at one locus is generated based on information at other loci.}
\label{Fig5}
\end{figure}

A challenge to this method arises when using the uniform MAX-3-SAT instances because, by the time fitnesses in the population increase substantially, the genomes are already similar to each other, making the shuffled genomes similar to the original ones. Therefore, for a clearer comparison, we used problems with distinct and varied solutions based on even-parity Boolean functions with $2$--$5$ variables (see Section \ref{subsection_problems}). Although parity functions are biologically unrealistic because they exhibit no sense of modularity, we use them here not to demonstrate the ability of the RBM-based model but to demonstrate that it takes into account interactions between genes in the generation of genetic information. For these parity function problems, let vectors with parity 1 have a fitness value of $1$ and let all other vectors have a fitness value of $0$. Note that the final bit of a correct solution is determined by the first $n-1$ bits of the solution. For example, in the $3$-variable problem, if the first two bits are $(1,0)$, the last bit must be $1$. In the simulations, we let the RBM complete the last bit of each of $100$ vectors for even-parity Boolean functions of different sizes. Figure ~\ref{Fig5} illustrates the results for these parity problems across various population sizes when the known gene values are sampled from the population.

The other way to demonstrate that genes interact in generating genetic information is to examine the internal workings of the RBM rather than treating it as a black box. Since biases represent separate effects of individual genes and connection weights represent genetic interactions, by comparing the impact of weights and biases, one can determine whether the RBM is able to recognize interactions between genes. For this purpose, we examine $\bar{w}$ in three different models: (1) the original model with both weights and biases; (2) a weights-only model where biases are set to $0$ and never updated; and (3) a biases-only model where weights are set to $0$ and never updated.

\begin{figure}
\centering 
\includegraphics[trim=2.6cm 7.0cm 3.1cm 10.1cm, clip=true, scale=0.9]{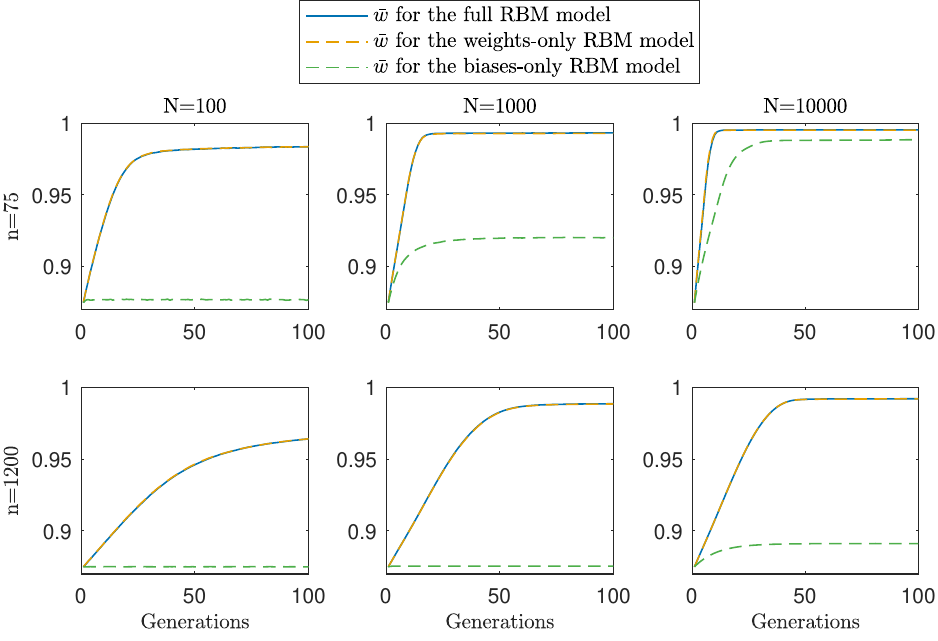}
\caption{$\bar{w}$ over the generations for two MAX-$3$-SAT instances ($75$ and $1200$ variables) averaged over 100 independent runs for three different models: the full model (solid blue line), the weights-only model (dashed orange line) and the biases-only model (dashed green line). Results are shown for three different population sizes: $N=100$ (left), $N=1000$ (middle) and $N=10,000$ (right). The other RBM parameters are $B = 10$, $S_{RBM} = 50\%$, $T = 20$, $H = n$ for both $n$ values; $\eta = 0.01$ for $n=75$ and $\eta = 0.001$ for $n=1200$. In the 75 variables case, for bigger population sizes, both biases and interactions in the generation of information in the genome contribute to fitness. In the 1200 variables case, the importance of interactions entirely dominates that of the biases.}
\label{Fig6}
\end{figure}

Figure~\ref{Fig6} demonstrates results of these models for uniform random MAX-$3$-SAT instances with $75$ (first row) and $1200$ (second row) variables and different population sizes. We can see that for all considered combinations of parameters, the full and the weights-only models yield similar results, while the biases-only model yields substantially lower population mean fitnesses.

\subsection{The relation between selection and the amount of genetic variation produced}

According to standard evolutionary theory, an adaptation could hypothetically evolve that increases the average mutation rate when the individual senses stress \cite{Radman1999,RamHadany2012,RamHadany2014}. Such an adaptation is considered to be an addition to rm/ns that could speed up the search for solutions by rm/ns when needed. However, there is no direct relationship between the direct pressure of natural selection on the population (differential survival and reproduction \textit{per se}, as opposed to stress sensed by individuals) and the overall amount of genetic variation produced (as opposed to maintained). IBE, in contrast, is a different process, where uniformity across individuals evolves together with the adaptation \cite{Livnat2013,Livnat2017}.

Consistent with IBE, in the simulations above, the amount of genetic variation produced is automatically reduced over time as the population closes in on the solution to the adaptational problem. Uniformity between individuals arises together with improvement, as the internal accumulation of information modeling the outside world is restricting further mutations.

\begin{figure}
\centering 
\includegraphics[trim=3.4cm 7.0cm 3.9cm 8.3cm, clip=true, scale=0.9]{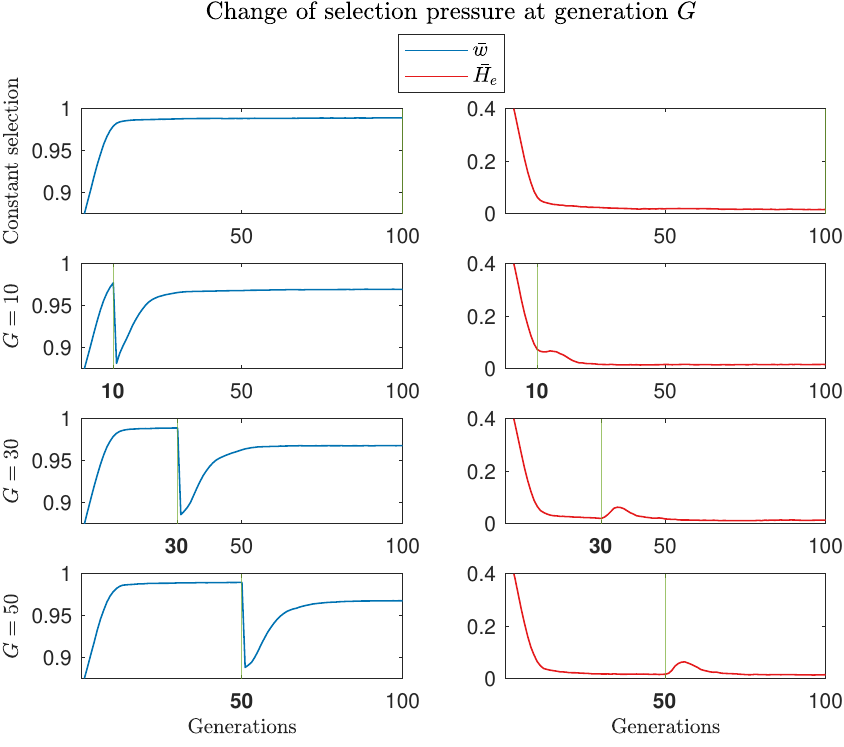}
\caption{$\bar{w}$ over time (left, blue) and the average expected heterozygosity, $\bar{H_e}$ over time (right, red) in the RBM model, averaged over $100$ independent runs, for MAX-3-SAT instances with $n = 20$ variables. Let $H_e = 1 - p^2 - q^2$, where $p$ and $q$ are frequencies of the two alleles, $0$ and $1$, in a given locus, with $p+q=1$, and let $\bar{H_e}$ be the mean $H_e$ across loci, serving here as a measure of genetic variation. The top two panels show the results of simulations where the same selection pressure is maintained from beginning to end. All other rows show the results of simulations where the selection pressure is changed at generation $G$ from one MAX-3-SAT instance to another ($G=10$, second row; $G=30$, third row; and $G=50$, bottom row). Other parameters are $N = 100$, $B = 10$, $T = 20$, $H = n$, $\eta = 0.1$, and $S_{RBM} = 50\%$.}
\label{Fig7}
\end{figure}

Additionally, we tested whether the amount of genetic variation produced can be increased by changing the selective regime at different generations. As shown in Figure~\ref{Fig7}, following a change in natural selection, the amount of variation produced first increases and then decreases again as improvement comes. Notice that, although the change in the amount of variation produced may seem small, it is only small in comparison to the amount of variation in the first generation, which is artificial to begin with.\footnote{In nature, there is no ``generation zero'' that begins with a maximal amount of variation generated at random.}

\subsection{The bell-shaped distribution of continuous traits and the maintenance of variation under selection}

\begin{figure}
\centering 
\includegraphics[trim=0.0cm 11.0cm 3.0cm 11.2cm, clip=true, scale=0.87]{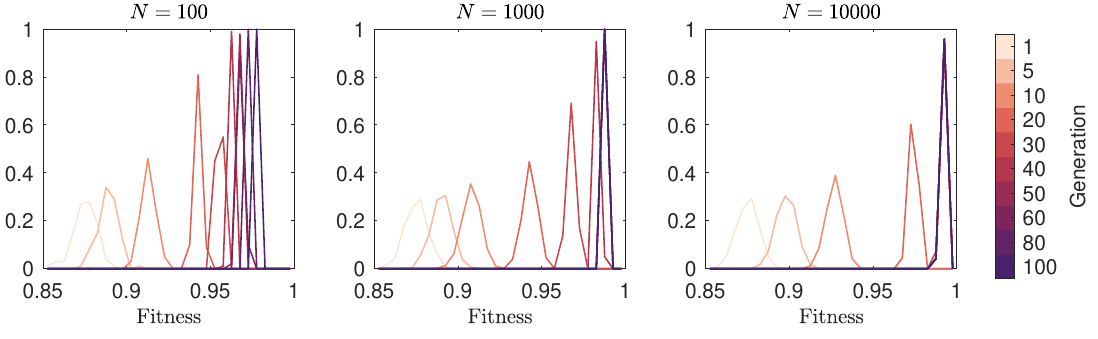}
\caption{Distributions of the fitness values in the population at different generations in the RBM model for a MAX-3-SAT instance with $n=600$. Fitness values were distributed across 0.005-sized bins and the number of fitness values per bin was normalized by the population size, $N$. Different panels show results for different population sizes ($N=100$, left; $N=1000$, middle; and $N=10000$, right), and, in each panel, fitness distributions for different generations between $1$ and $100$ are shown based on the color chart. The parameters were selected to provide the highest tested $\bar{w}$ as measured at generation $100$. Specifically, for $N=100$: $B=10$, $S_{RBM}=50\%$, $H=2n$, $T=20$ and $\eta=0.01$. For $N=1000$: $B=10$, $S_{RBM}=50\%$, $H=2n$, $T=100$ and $\eta=0.001$. For $N=10000$: $B=100$, $S_{RBM}=10\%$, $H=n$, $T=100$ and $\eta=0.01$.}
\label{Fig8}
\end{figure}

The bell-shaped distributions commonly observed in nature for continuous traits like height are attributed in standard evolutionary theory to additive functions of alleles across loci in accord with the Central Limit Theorem (CLT)  \cite{Fisher1918,FalconerMackay1996}. In our simulation of the MAX-SAT adaptational challenge, the variables interact in a complex manner, and therefore selection acts on genomes as complex wholes; yet the distribution of fitness values in the population is also bell-shaped. In addition, the range of phenotypic variation is maintained for multiple generations of selection. Figure~\ref{Fig8} shows the distributions of fitness values for different generations in the RBM model for a MAX-3-SAT instance with $n=600$ variables. The distributions are approximately bell-shaped and the mean increases over the generations. Thus, neither mutational randomness nor additivity of independent fitness contributions are necessary for the emergence of bell shaped distributions or for the increase in $\bar{w}$.

\section{Discussion}

It is not surprising that the RBM-based model outperforms the random mutation one given previous results in EDAs \cite{TangShimTanChia2010,ShimTanChia2010,ShimTan2012,ShimTanChiaMamun2013,Shim_etal2013,Ceberio_etal2014,Liang_etal2015,Probst_etal2017,BaoSunGongZhang2022}. Rather, our goal has been to point out that multivariate EDAs instantiate a third category of heritable change: nonrandom, non-Lamarckian mutation. Indeed, such EDAs enable interactions between genes in the generation of heritable information. This shows that multivariate EDAs are natural in an important sense, clarifies that the IBE concept of nonrandom, non-Lamarckian mutation is logically coherent, and points at a new direction for theoretical research, as we will explain below.

The results demonstrate that heritable change can be neither random nor Lamarckian as follows: It is not random because it depends on information in a biologically meaningful manner. It is not Lamarckian because the information is internal and is accumulated in the genome over generations. In stark contrast with Lamarckism, selection remains the only source of feedback on the fit between the organism and its environment. In stark contrast with the notion of random mutation, mutation is fundamentally nonrandom.

At the same time, the results demonstrate that the mechanisms of nonrandom, non-Lamarckian mutation can be natural in an important sense. As an example, an analogy exists between the Hebbian learning mechanism of adjustment of connection weights in Boltzmann Machines and the biological gene-fusion mutational mechanism described by Bolotin et al. \cite{Bolotin_etal2023}. In accord with the former, units that ``fire together wire together'' \cite{Hebb1949,LowelSinger1992}, and in accord with the latter, genes that are used together are more likely to become fused together by mutational mechanisms \cite{Bolotin_etal2023,LivnatMelamed2023}. Indeed, there is no ``homunculus'' inside the individual deciding on heritable changes in a centralized manner: mutations in a given gene depend on the genes interacting with that gene. Heritable change is local and influenced by a fan-in of information, allowing for decentralized integration of information over the generations.

Despite its differences from IBE, the model already captures certain aspects of IBE indirectly:
\textit{a}) Sexual recombination, selection, and nonrandom, non-Lamarckian mutation interact in a complementary fashion \cite{Livnat2013}. 
\textit{b}) Evolution is driven by an interaction of simplification and fit \cite{Livnat2017}.  
\textit{c}) Selection on individuals as complex wholes is at the core of the adaptive evolutionary process \cite{Livnat2013,Livnat2017}. 
\textit{d}) Randomness does not generate improvements directly but rather enables generalization by the manner in which it is connected with the rest of the evolutionary process \cite{Vasylenko_etal2020}. 
\textit{e}) Evolution takes place at the level of the population as a whole: an improvement arises not from a random mutation that originated in one individual and spread to the entire population, but from internal 
simplification mechanisms that naturally integrate information over generations under selection from many surviving individuals \cite{Livnat2013,Livnat2017}. 

Despite its lack of realism, the model offers a new angle on difficult biological problems. First, that improvement comes together with uniformity between individuals offers a theoretical explanation for observations suggesting that selection \textit{per se} can increase the production of genetic variation \cite{Darwin1859}. Second, while it has been assumed that additive genetic effects explain the commonly observed bell-shaped distribution of continuous traits \cite{Fisher1918,FalconerMackay1996}, the results suggest that a bell-shaped distribution can also be obtained from a complex function where alleles interact across loci, and that genetic variation in the direction of selection can be replenished by nonrandom, non-Lamarckian mutation.

These results point at a new direction for theoretical research: internal fan-in of information through heritable change needs to be incorporated both in evolutionary theory and in evolutionary computation. While genetic algorithms (GAs) apply concrete mechanisms of recombination and mutation, these mechanisms have not incorporated true, general, internal fan-in of evolving information; and while multivariate EDAs do have such integration of internal information, they instantiate it in a manner that is highly abstract and indirect from a biological perspective. Future research is needed to find ways to combine concrete, natural operators of recombination and mutation together with internal integration of information. Empirical observations of IBE \cite{LivnatMelamed2023,Bolotin_etal2023,Melamed_etal2025} can serve as a source of inspiration for modeling such mechanisms of nonrandom, non-Lamarckian heritable change explicitly. Below, we elaborate on connections to evolutionary theory, computational learning theory, and other areas.

\subsection{The reconciliation between Darwinism and Mendelism revisited}

The role of sexual recombination has been problematic both in evolutionary theory and in evolutionary computation. The layperson’s intuition has been that sexual recombination generates a vast number of different combinations of genes, and that, because variation adds ``fuel’’ for selection, this must speed up evolution. However, sexual recombination cannot make selection favor the best particular combinations among many because it constantly breaks down the combinations it generates  \cite{EshelFeldman1970,Lewontin1971,BartonCharlesworth1998,LivnatPapadimitriou2016alg}. Therefore, standard evolutionary theory deviated from the layperson's intuition in search of solutions of other types.\footnote{As noted, in such solutions, alleles at different genes are selected separately \cite{Fisher1930,Muller1932}, or combinations of alleles at different loci play a role but only in a narrow sense (e.g., refs. \citenum{Hamilton1980, Jaenike1978}).} However, each such proposed solution required its own special conditions \cite{West_etal1999}, in contrast to the prevalence of sex in nature, and all treated sexual recombination as an auxiliary process.\footnote{20\textsuperscript{th}-century theories on the role of sex in evolution have considered it a ``bonus’’ that may speed up evolution under certain conditions but is not essential for evolution. As H. J. Muller wrote, ``There is no basic biological reason why reproduction, variation and evolution can not go on indefinitely without sexuality or sex; therefore, sex is not, in an absolute sense, a necessity, it is a `luxury'’’ \cite{Muller1932}.} Relatedly, in the field of genetic algorithms (GAs), the problem of the breaking down of favorable combinations has been noted \cite{Goldberg_etal1989,GoldbergBridges1990,Pelikan_etal2002}, even while it has been simultaneously assumed that recombination and selection should generate adaptive complexes \cite{Holland1975,DeJong1975,Goldberg1989}.

To move beyond this conceptual state of affairs, one must trace its roots back to the fundamental problems in Darwin's theory and how it was reconciled with Mendelism. In Darwin's theory, adaptive evolution resulted from selection acting on small but abundant variation existing among individuals in every generation, and proceeded gradually \cite{Provine1971}. Two fundamental problems with this theory were the idea that sexual reproduction would obliterate any heritable variation and prevent this variation from being favored by selection over generations (``blending inheritance''), and ignorance of how heritable variations arise \cite{Provine1971}. In a seeming contrast, Mendel's finding that the shuffling of heritable factors (later called ``genes'') through sexual recombination determined discrete traits such as the color of the peas (which could be either yellow or green) suggested that heritable information was discrete (``particulate'') \cite{Mendel1866} and seemed to support the view that evolution progressed in discrete mutational leaps \cite{Provine1971}. Fisher proposed to reconcile Darwinian gradualism with Mendelian particulate inheritance by assuming that multiple genes can make separate, additive contributions to the same continuous trait \cite{Fisher1918}. This assumption allowed beneficial alleles of discrete genes to spread simultaneously in a sexual population without disrupting each other \cite{Fisher1930,Muller1932}, and thus allowed continuous traits to evolve gradually over the generations under natural selection.At the same time, he assumed mutations to be accidental \cite{Fisher1930}, avoiding Lamarckism in a manner which resonated with the view of genes as separate actors.

This solution, however, which became foundational to 20\textsuperscript{th}-century evolutionary theory, has left fundamental problems open. First, it offered a framework under which sexual recombination was not disruptive---which is not the same as explaining why a trait of such fundamental importance exists in the first place and what role it plays in evolution \cite{Livnat2013,LivnatPapadimitriou2016alg}. Second, it offered a model where a single numerical variable called ``fitness'' could additively increase over generations based on independent genetic contributions\footnote{When genetic interactions affect the fitness in focus, a general increase in fitness in a sexual population cannot be properly demonstrated within the traditional framework \cite{Ewens2004,Okasha2008}.} \cite{Fisher1930}---which is not the same as showing how complex structure evolves \cite{Valiant2009}. Indeed, the underlying assumption that genes make separate contributions to fitness has been criticized as paradoxical from a biological standpoint \cite{Wright1930,Mayr1955,Mayr1963}.

IBE offers a different reconciliation based on recognizing the interaction of two forces in evolution—an internal and an external one \cite{Livnat2013,Livnat2017,Melamed_etal2022,LivnatMelamed2023,Bolotin_etal2023,Melamed_etal2025}. Under this framework, as demonstrated here indirectly, sexual recombination generates individual genetic combinations as complex wholes, selection acts on these combinations as complex wholes, and nonrandom, non-Lamarckian heritable changes enable these complex wholes to have effects on future generations despite their transient nature, based on fan-in of information from multiple recombining loci \cite{Livnat2013}. Thus, sexual recombination, natural selection and the empirical nature of heritable change are connected in a manner that captures the central role of interactions between genes.

\subsection{Heritable change integrates internal information}

As argued by IBE and demonstrated here using the multivariate EDA framework, the problem of sexual recombination is intrinsically linked to the problem of heritable change. In standard evolutionary theory, mutation is accidental: local change in the genome occurs in a manner unrelated to information at other genomic loci. In contrast, IBE holds that there \textit{is} fan-in of information in the origination of heritable change. This changes how the fundamental elements of evolution---recombination, selection and mutation---interact. In standard theory, separate pieces of information inherited from different genomes merely sit together side by side transiently in the same genome until they are separated again, and therefore the breaking down of favorable combinations is a problem. In contrast, in our model, such pieces of information are actually combined and integrated through heritable change. Therefore, fit complex wholes have effects on future generations through these changes, and the constant breaking down of favorable combinations is not a problem---it is how things work.

As is well known, putting together pieces of information is at the basis of computation, as seen in the operation of a Turing machine, a logical gate or a neuron. We claim that it is also at the basis of evolution (see also \citenum{Livnat2013,LivnatPapadimitriou2016TREE}), although using it here requires a conceptual change in what roles mutation and selection play and how they interact,\footnote{In addition, note that, more precisely, heritable changes parallel learning changes (such as Hebbian learning, which also requires fan-in from different neurons)---i.e., changes in the structure of the network, not computations that are a part of its day-to-day operation.} as discussed below.

\subsection{Parsimony and fit in biological evolution}

The action of the RBM can be seen as demonstrating local regulatory simplification: the initially complex influences of the network on the correlation between the activations of a pair of units are replaced over time by the strengthening of a single, direct connection between these units. The principle of regulatory simplification accords not only with the used-together-fused-together gene fusion mechanism \cite{Bolotin_etal2023}: it has been argued that mutations of various types, including point mutations, deletions, translocations and fusions, naturally simplify and replace preexisting genetic interactions\cite{LivnatMelamed2023,Melamed_etal2025}.

Indeed, that evolution is driven by an interaction between an external force of natural selection and an internal force of natural simplification, and that this interaction uncovers the commonalities between successful individuals, is a key principle of IBE \cite{Livnat2013, Livnat2017}; and this principle is also seen in operation in the simulation. Because of selection, the information that is processed by the RBM is composed of genomes that survived and reproduced; and because of its simplifying action, the RBM identifies patterns of genetic interactions that are common among successful individuals.

\subsection{Changed conditions and the amount of heritable variation produced}

In the \textit{Origin of Species} \cite{Darwin1859}, Darwin emphasized that changed conditions lead to an increased production of heritable variation, as in the case of domestication (e.g., ref. \citenum{Darwin1859}, Ch. 1). These observations were largely ignored in the 20\textsuperscript{th} century despite supportive empirical data \cite{Nevo_etal1984} and despite the lack of a straightforward explanation otherwise \cite{Leffler_etal2012}.

The results observed here that the rate of production of variation is higher under changing conditions is consistent with IBE as follows. \textit{a}) Because mutation origination depends on information in the genome, the amount of variation produced increases with the amount of variation present. \textit{b}) The closer the population is to a solution, the fewer ways there are to be as close to that solution. From (\textit{a}) and (\textit{b}) together, as the population closes in on an adaptation, less variation is produced that is relevant to this adaptation. In addition, the present selection pressure is also important for generating variation: without a coherent selection pressure, no regularities among successful individuals can be found that allow the generation of meaningful heritable changes as modelled here. These mechanistic considerations are fundamentally different from hypotheses in evolutionary theory whereby an adaptive trait that has evolved by rm/ns makes individuals increase the production of heritable variation in response to stress that they sense within their lifetime \cite{Radman1999,RamHadany2012,RamHadany2014}. Here, it is differential survival \textit{per se} that affects the production of variation through nonrandom, non-Lamarckian mutation, not the sensing of stress by individuals.

\subsection{Bell-shaped curves and the maintenance of variation in the direction of selection}

Fisher argued that the bell-shaped distributions often observed in nature for continuous traits like height are explained by the assumption of additive contributions of separate genes to the same trait, in line with the CLT \cite{Fisher1918}. However, we argue that the summation of independent random variables is not the only possible explanation for bell-shaped distributions: complex functions that are not dominated by any small subset of their  variables may generate the observed distributions as well, as in our simulation.

Also of interest in this connection, in accord with Beatty's analysis of the history of evolutionary biology \cite{Beatty2016,Beatty2019}, is that in our simulations, the distribution slides along with its mean with little effect on its standard deviation for multiple generations before the mean begins to plateau (Figure~\ref{Fig8}). This point is of interest because it bears on the conceptual reasons for the assumption of random mutation to begin with in \cite{Beatty2016}. Darwin relied on the existence of ever-present small heritable variation in all directions to enable selection to lead the way in evolution \cite{Provine1971,Beatty2016}. However, when selection consumes heritable variation in a particular direction, what would replenish variation in that direction specifically and thus enable the accumulation of many small selected changes in the same direction in the long term? Morgan considered that, by implying such replenishing in a specific direction, the assumption of ever-present variation in all directions introduced purpose into the evolutionary process \cite{Beatty2016} and, precisely in order to avoid this conclusion, he emphasized that there must be no connection between selection and the causes of mutation \cite{Morgan1925,Morgan1935,Allen1978,Beatty2016}---that mutation must be random---and that it was not possible for phenotypic variation to be consistently maintained in the direction of selection.\footnote{Morgan argued that such variation must eventually disappear until a new, rare, accidental mutation in the relevant direction arises and that, had there been a mechanism whereby survivors produced further variation in the direction of selection, it would have provided \textit{the} explanation for the evolution of life, but that such a mechanism did not exist \cite{Morgan1935,Beatty2016}.} In Beatty's analysis, during the formation of the modern synthesis, investigators maintained Morgan's separation between natural selection and the origin of genetic variation, but suggested how variation could be maintained after all in the direction of selection: by interaction with alleles that have spread under selection, alleles of other genes previously neutral to the selection pressure could become relevant to it, allowing selection to generate its own variation \cite{Chetverikov1926,Beatty2019}. This particular modern-synthetic argument, however, is inconsistent with the rest of the synthesis: separate, additive effects across genes were assumed to explain the bell-shaped distribution of continuous traits and to enable effective selection on randomly produced variation despite sexual recombination, even while interactions between genes and sexual recombination were assumed to enable the range of variation to be indefinitely maintained over generations. In contrast, our model offers a different and self-consistent view: complex genetic interactions are responsible both for the observed bell-shaped distributions of continuous traits and for the maintenance of variation in the direction of selection. Selection maintains variation in its own direction through nonrandom, non-Lamarckian mutations, which require interactions between genes.

\subsection{The role of randomness}

According to standard evolutionary theory, the accident that supposedly generates a mutation endows that mutation with its own effect on the organism \cite{Morgan1903,Fisher1930,Dawkins1986,Futuyma1998}, and it only remains for selection to test whether this effect is beneficial or not, as though random bits encode novel adaptive information directly. In contrast, in our simulation, random bits in and of themselves do not encode new adaptive information. Instead, they enable generalization by means of how they are connected with the rest of the evolutionary process: First, the RBM learns those relationships between alleles at different genomic loci that are common among surviving individuals. Next, the space of possible genotypes defined by the rules thus learned is randomly sampled by the free-running state of the RBM.

This use of randomness relates to mixability theory  \cite{Vasylenko_etal2019,Vasylenko_etal2020}, better resonates with how randomness has been used in powerful algorithms \cite{MotwaniRaghavan1995,Wigderson2019}, and contrasts with traditional hill-climbing analogies in evolution \cite{Wright1932,KauffmanLevin1987,Dawkins1996,Gavrilets2004}. In the latter, the life form evolves step by step by a sequence of random mutations from a low-fitness to a high-fitness point. In contrast, in our simulation, randomization events of consecutive generations in and of themselves do not add up to form a direction. Instead, the randomization that takes place at each generation expands on what already exists, and improvement arises from gradual convergence on a solution over generations\footnote{The part of the population favored in one generation is ``expanded,’’ the part of this expansion favored in the next generation is expanded, etc.} \cite{Livnat2013}.

\subsection{Evolution as learning}

Several lines of work have attempted to explore theoretically the connection between evolution and learning (e.g., \citenum{Feldman2008,Valiant2009,Valiant2013,Chastain_etal2014,Watson_etal2014,AngelinoKanade2014,WatsonSzathmary2016,LivnatPapadimitriou2016TREE,Livnat2017,Kouvaris_etal2017,Vanchurin_etal2022}). Among them was Valiant's pioneering work on the evolvability model \cite{Valiant2009,Valiant2013}. That work  set to answer the question of whether rm/ns could account for the evolution of life given the amount of time that has been available for it \cite{Valiant2009}. Analytical population genetic models could not answer this question because, while they address the evolutionary dynamics of allele frequencies given selection coefficients etc., they do not include a representation of complex biological structure. Using the Probably Approximately Correct (PAC)--learning framework \cite{Valiant1984}, Valiant constructed an analytical model with a representation of the phenotype in order to relate kinds of adaptations (functions, in the model) to the time it would take them to evolve by rm/ns, with all resources being polynomial in the number of inputs that may be required in the adaptive mechanism and in the inverse of the evolved error that would be acceptable in principle \cite{Valiant2009}.

In Valiant's model, in the course of evolution, among a total set of possible literals and their negations, the correct subset of literals must be found that forms the domain of a given, many-arguments Boolean function such as conjunction, disjunction or parity \cite{Valiant2009}. The evolving ``hypothesis'' is a set of literals, which is updated with a minimal ``population'' as in the ``(1+1)'' evolutionary algorithm \cite{BackSchwefel1993,Droste_etal2002}: a single individual, representing the evolving hypothesis, generates one more individual by a mutation that at random either adds a certain literal, drops a certain literal, or swaps one literal for another \cite{Valiant2009}. New individuals are produced such that either a beneficial mutation is fixed or, if no such mutation exists, a slightly neutral random mutation is fixed.

However, in developing the evolvability model, Valiant followed the dichotomy of random mutation vs. Lamarckism \cite{Futuyma1998}: he argued that the evolutionary algorithm cannot understand the highly complex relationship between genotype and phenotype, and therefore evolutionary change cannot meaningfully depend on information regarding the situations that the individual encounters during its lifetime and the correct responses to them, but can only depend on the individual's overall fitness. \footnote{Considering “examples” to be input vectors received from the environment to which the individual needs to respond appropriately, and different individuals to be “competing hypotheses,” Valiant wrote: “In evolution, we assume that the updates depend only on the aggregate performance of the competing hypotheses on a distribution of examples or experiences, and in no additional way on the syntactic descriptions of the examples. This restriction reflects the idea that the relationship between the genotype and phenotype may be extremely complicated, and the evolution algorithm does not understand it.” (\citenum{Valiant2009}, p. 3)} This exclusion of dependency on information from the environment is the exclusion of Lamarckism, and, in the evolvability framework, it turns evolution into a restricted form of the Statistical Query (SQ) model \cite{Kearns1998}---itself a restricted form of standard PAC learning \cite{Valiant1984}. With this exclusion, Valiant obtained that Boolean monotone conjunctions and disjunctions are evolvable\footnote{``Evolvability'' means that for a given such type of function, the set of relevant literals can be found by a process meant to capture random mutation and natural selection that is efficient in terms of the time and space resources used.} with polynomially bounded resources from any starting hypothesis over a uniform distribution of inputs, whereas Boolean parity functions, which are unnatural, are not \cite{Valiant2009}. If one intuitively takes this restricted form of PAC learning to be weak, one could infer either that a weak biological evolutionary process is sufficient to produce life as we know it, or that the model does not capture real-world biological evolution to begin with \cite{Davis2013}. However, according to IBE, there \textit{is} a way for the evolutionary process to depend meaningfully not only on the fitnesses of individuals, and not by Lamarckism; that is, mutations are neither random nor Lamarckian: the change of a variable depends on the values of other variables.

To see how the IBE view is possible using the evolvability model as a starting point, one needs to rearrange its elements. In the evolvability model, the individual is the ``hypothesis’’ that is evolving, and the inputs that the individual encounters during its lifetime are the ``examples’’ that the hypothesis needs to classify correctly. In contrast, in our model, the individuals are the examples, whereas the hypothesis is at the population level and is about what makes these examples ``positive’’---what makes individuals survive and reproduce. The hypothesis is used to produce examples, selection chooses the better examples among these, and the chosen examples serve as the basis for hypothesis improvement. Thus, unlike standard PAC learning, the examples used in the next generation are drawn neither independently nor from a distribution that is fixed across generations: as the hypothesis evolves, the example distribution changes. This arrangement represents assumptions that are weaker not only than those of SQ but also than those of PAC without queries.

Although Valiant allowed mutation to carry out any process that is polynomially computable by a Turing Machine so that one could account for duplications and deletions of sequences as well as for point mutations, as long as the evolvability framework does not explicitly suggest mutations that depend on genetic content and fan-in, all mutation types in it will remain random in a crucial sense. The attempts in genetic algorithms to use mutation types such as inversions that were independent of the genetic content followed the same approach, sans 
internal integration of information \cite{Holland1975,GoldbergBridges1990}.
In contrast, we argue that the full consequences of the Turing completeness of mutation need to be explored in the case where evolving information is integrated by heritable change---namely within the IBE conceptual framework \cite{Livnat2013,Livnat2017,LivnatMelamed2023}.

Reconsideration of terminology such as ``genetic algorithms'' and ``learning'' can now sharpen the connection between evolution and cognition. Although Valiant cast evolution in terms of PAC learning and argued that this offers a unifying framework for evolution and cognition, in the evolvability model, evolution is actually a restricted form of PAC that is akin to basic EAs and univariate EDAs. In contrast, it is when mutations are neither random nor Lamarckian that mutational mechanisms become analogous to mechanisms of cognition and learning by the brain, such as Hebbian learning \cite{Hebb1949,LowelSinger1992} and chunking \cite{Lindley1966,TulvingCraik2005}. Thus, the modeling of mutations with fan-in of information brings evolution and cognition much closer together.

\subsection{Future research}

About a century ago, when Fisher attempted to provide a theory for evolution based on random mutation, it would have been incomparably harder to incorporate structure representing nonrandom mutation. Yet even today, standard population genetics, basic EAs, univariate EDAs, and the evolvability model based on standard PAC are all easier to analyze than multivariate EDAs \cite{LehreNguyen2019,KrejcaWitt2019,LarranagaBielza2024}, PAC with queries \cite{Diakonikolas_etal2024}, and IBE. As Wigderson noted, standard PAC ``sweeps under the rug'' the fact that ``in nature and practice, the hypothesis of the learner often generates action/behavior that affects the environment and future examples it may generate'' (\citenum{Wigderson2019}, p. 249).

Interpreting multivariate EDAs through an IBE lens highlights a lacuna in previous research: neither in evolutionary theory nor in standard GAs has internal fan-in of information through heritable change been considered. In standard GAs, operators are applied at random, not guided by internal genetic information. In multivariate EDAs, internal integration of information is employed, but only in a biologically unrealistic manner. Future work should employ natural, concrete mechanisms of recombination and mutation as in GAs, but with internal fan-in of information as in multivariate EDAs. For this purpose, empirical observations within the IBE framework can be drawn upon \cite{LivnatMelamed2023,Bolotin_etal2023,Melamed_etal2025}.

Examining the Turing completeness of heritable change as an internal integrator of evolving information is not a minor conceptual change but requires a reconceptualization of how evolution happens. Using the evolvability model as an example, it requires a reconceptualization of what counts as hypothesis and examples, what roles the fundamental elements of evolution play and how they interact with each other---including that evolution is driven by the interaction of parsimony and fit.

In pursuing this new direction, additional considerations should be kept in mind. First, in contrast to the RBM mechanism, which itself does not evolve over the generations, IBE argues that no arbitrary line separates between a fixed apparatus causing mutations on one hand and an ever-changing part of the genome on the other. Allowing mutational mechanisms to evolve poses a challenge for future research.

Second, although beginning with randomized genomes and ending at equilibrium, as in our simulation, is a ubiquitous approach in evolutionary computation, this approach is biologically unrealistic. In reality, biological evolution neither starts at a tabula-rasa state nor stops. Diversity keeps being generated and new adaptations keep arising, not out of thin air but from the coming together of preexisting, simplified, well-working elements \cite{Livnat2017}.\footnote{In fact, evolution itself is not known to have arisen from random noise at a particular point in space and time. Instead, there could have been a gradual transition from a ``chemical’’ world to a world more and more ``biological’’ \cite{Livnat2013}.} Although we have not attempted to solve these deep problems here, one must keep them in mind when interpreting the results. For example, while in our simulation, the amount of relevant variation decreases in time, leading the RBM process to plateau, in nature, new variation and adaptations constantly arise.

In addition to these challenging directions for future research, even the rudimentary model presented here can be explored further. Early evolution in the random mutation model seems to be finding the ``low-hanging fruit’’---the special solution-bits that satisfy many clauses (Nick Pippenger, personal communications), perhaps conceptually related to the basic approximation algorithm for \textit{k}-MAXGSAT  \cite{Papadimitriou1994}. In contrast, one of the most surprising features in the results is the rapid transition from low to high fitness of the RBM model in the transient, which raises several questions (Nick Pippenger, personal communications): What kind of learning is it doing? What is gained in the early generations that leads to a sudden transition in the transient? And, by the time the process ends, has the population learned in some sense the fitness function?

\subsection{Conclusions}

The RBM-based multivariate EDA model captures in an abstract and indirect manner several aspects of IBE: 
\textit{a}) Heritable change is neither random nor Lamarckian: it responds to information accumulated in the genome over generations. At each generation, heritable changes arise based on the information accumulated internally thus far, and those that survive become a part of that information, affecting future generations’ heritable changes. 
\textit {b}) Recombining pieces of heritable information do not merely sit side by side until they are separated again, but are actually integrated through heritable change. As a result, the basic elements of evolution interact: sexual recombination generates a large variety of different combinations of genes; selection acts on these combinations as complex wholes; and these combinations have an effect on future generations through the mutations derived from them, enabled by mutational fan-in of information. 
\textit {c}) The evolutionary process simulated is driven by the interaction of parsimony and fit: while the external force of natural selection provides the feedback on the fit between the organism and its environment, an internal force of natural simplification, driven by nonrandom, non-Lamarckian heritable change, integrates information, finding the commonalities between successful individuals---the principles underlying success. 
\textit{d}) Randomization does not directly invent improvements but enables generalization by the manner in which it is connected with the rest of the evolutionary system. 
While demonstrating the above, the results show a connection between the strength of selection and the generation of genetic variation, the ability of a complex function involving interactions of alleles across loci to generate a bell-shaped distribution of a trait value, and the ability of this distribution to follow the direction of selection. Given the importance of sufficiently large population sizes in nature, it is also of interest that the RBM model requires a sufficiently large surviving population to learn from, whereas the RM model is encumbered by increasing the number of surviving individuals much beyond the top performers.

The basic requirements on the nature of mutation in this model are natural and fitting in principle with a large number of empirical observations. This fact demonstrates concretely that another way of thinking is possible about how mutations arise and how evolution happens at the fundamental level besides rm/ns and Lamarckism. Future work will be needed to attempt to model IBE more realistically, one option being to combine the naturalness of concrete recombination and mutation mechanisms as in GAs with the fan-in of information through heritable change as in IBE.

\section*{Competing interests statement}

The authors declare no competing interests.

\section*{Acknowledgments}

We thank Sasha Bolshoy, Amit Livnat, Mor Naftali, Nick Pippenger, Yuri Rabinovich and Kim Weaver for extensive help and comments. This publication was made possible through the support of the Sagol Network, ISF grant 1986/16 to A.L., and John Templeton Foundation grant 62220 (subaward) to A.L. Computations presented in this work were partly performed on the Hive computer cluster at the University of Haifa, which is partly funded by ISF grant 2155/15. The opinions expressed in this publication are those of the authors and do not necessarily reflect the views of the John Templeton Foundation.

\newpage
\clearpage

\bibliographystyle{pnas-new}

\end{document}